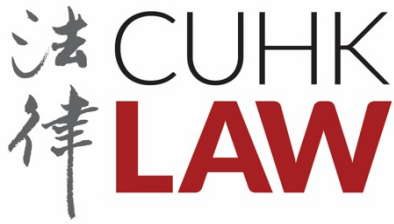

THE CHINESE UNIVERSITY OF HONG KONG

FACULTY OF LAW

Research Paper No. 2024-4

# Caveat Lector:

# Large Language Models in Legal Practice

Eliza MIK

**Caveat Lector: Large Language Models in Legal Practice**

**Eliza Mik***

## Abstract

Reader Beware. The current fascination with large language models, or "LLMs," derives from the fact that many users lack the expertise to evaluate the quality of the generated text. LLMs may therefore *appear* more capable than they actually are. The dangerous combination of fluency and superficial plausibility leads to the temptation to trust the generated text and creates the risk of overreliance. Who wouldn't trust perfect legalese?

Drawing from recent findings in both technical and legal scholarship, this Article counterbalances the overly optimistic predictions as to the role of LLMs in legal practice. Integrating LLMs into legal workstreams without a better comprehension of their limitations, will create inefficiencies if not outright risks. Notwithstanding their unprecedented ability to generate text, LLMs do not understand text. Without the ability to understand the meaning of words, LLMs will remain unable to use language, to acquire knowledge and to perform complex reasoning tasks.

Trained to model language on the basis of stochastic word predictions, LLMs cannot distinguish fact from fiction. Their "knowledge" of the law is limited to word strings memorized in their parameters. It is also often incomplete and largely incorrect. LLMs operate at the level of word distributions, not at the level of verified facts. The operating propensity to hallucinate, to produce statements that are incorrect but appear helpful and relevant, is alarming in high-risk areas like legal services. For the time being, lawyers should beware of relying on text generated by LLMs.

## Table of Contents

---

\*          Assistant Professor, Faculty of Law, The Chinese University of Hong Kong (elizamik@cuhk.edu.hk). I thank Noam Noked, Stephen Gallagher, Jyh-An Lee, Dan Hunter and Shmuel I. Becher for their helpful comments. Research for this paper has been supported by a Direct Research Grant No 4059053 "Testing the Limits of Machine Learning Approaches in the Automation of Legal Tasks: From Legal Prediction and Reasoning to Document Management and Generation"

# Introduction

If one was to believe the mainstream predictions made less than 15 years ago, the legal profession should have been revolutionized by the introduction of IBM Watson, the uncanny AI system that beat human champions at the game of Jeopardy![1] If an AI could correctly answer hundreds of questions from different domains, it would surely be able to tackle legal problems! Instead, IBM Watson turned out to be a complete failure.[2] At present, it is commonly anticipated that the

---

[1] In a televised *Jeopardy!* contest viewed by millions in February 2011, IBM's Watson DeepQA computer made history by defeating the TV quiz show's two all-time champions, Brad Rutter and Ken Jennings, *see* https://www.ibm.com/history/watson-jeopardy. The DeepQA technology was marketed under the brand name Watson, *see*: David Ferrucci et al., *Building Watson: An Overview of the DeepQA Project*, 31 AI MAG 59-79 (2010).

[2] Gary Marcus, *Deep Learning is Hitting a Wall*, NAUTILUS (March 10, 2022); Eliza Strickland, *How IBM Watson Overpromised and Underdelivered on AI Healthcare*, (April 2, 2019) IEEE SPECTRUM, https://spectrum.ieee.org/how-ibm-watson-overpromised-and-underdelivered-on-ai-health-care; Steve Lohr, *What Ever Happened to IBM's Watson?* (Jul. 17. 2021) N.Y.TIMES
https://www.nytimes.com/2021/07/16/technology/what-happened-ibm-watson.html (The day after the Watson victory, IBM declared that they were "exploring ways to apply Watson skills to the rich, varied language of health care, finance, law and academia." The main scientist behind Watson, David Ferrucci, explained that Watson was engineered to identify word patterns and predict correct legal profession will be transformed by large language models, or "LLMs."[3] LLMs, which owe their name to the large number of parameters in their underlying neural networks and to the vast amounts of data they are trained on, generate text by predicting the likelihood of a token (character or word) given its preceding or surrounding context.[4] As their ability to generate plausible, human-like text is unprecedented, it is widely believed that LLMs will have a significant impact on the legal profession – be it by making lawyers more efficient or by reducing the need for lawyers altogether.[5] Such beliefs are, however, based on uninformed assumptions about the capabilities of LLMs and the nature of legal work. Equating legal work with the generation of text fails to acknowledge the reasons why lawyers "produce text," not to mention the expertise required to do so. Once the technical underpinnings of LLMs are examined, their transformative potential becomes less obvious – unless one subscribes to the reductive view that when performing most legal tasks, the generation of text can be divorced from understanding, reasoning, and knowledge. Although the initial enthusiasm surrounding LLMs seems to be subsiding,[6] most predictions about their significance for the legal profession continue to focus on their unprecedented capabilities rather than acknowledging their intrinsic shortcomings. This Article takes a more realistic approach and examines LLMs from the perspective of their limitations and from the perspective of their users. The technologies discussed below are complex, but the message is simple: for the foreseeable future, lawyers should beware of relying on any output generated by LLMs.

It must be remembered that many of the reported successes of LLMs are the result of cherry-picking[7] and that it is open to debate whether LLMs can perform

---

answers for the trivia game. It was not an all-purpose answer box ready for the commercial world and could fail a second-grade reading comprehension test.)

[3] *Generative AI could radically alter the practice of law*, THE ECONOMIST (June 6, 2023) https://www.economist.com/business/2023/06/06/generative-ai-could-radically-alter-the-practice-of-law; Joseph Briggs & Devesh Kodnani, *The Potentially Large Effect of Artificial Intelligence on Economic Growth*, GOLDMAN SACHS ECONOMICS RESEARCH (Mar. 23, 2023) https://www.key4biz.it/wp-content/uploads/2023/03/Global-Economics-Analyst_-The-Potentially-Large-Effects-of-Artificial-Intelligence-on-Economic-Growth-Briggs_Kodnani.pdf.

[4] Rishi Bommasani et al., *On the Opportunities and Risks of Foundation Models*, ARXIV at 3, 6–7, 49 (Aug. 18, 2021), https://arxiv.org/abs/2108.07258.

[5] Bommasani et al., *supra* note 4, 66.

[6] Daron Acemoglu, *Get Ready for the Great AI Disappointment*, WIRED (Jan. 10, 2024) https://www.wired.com/story/get-ready-for-the-great-ai-disappointment/ ("More and more evidence will emerge that generative AI and large language models provide false information and are prone to hallucination—where an AI simply makes stuff up, and gets it wrong. Hopes of a quick fix to the hallucination problem via supervised learning, where these models are taught to stay away from questionable sources or statements, will prove optimistic at best. Because the architecture of these models is based on predicting the next word or words in a sequence, it will prove exceedingly difficult to have the predictions be anchored to known truths.")

[7] Yonathan A. Arbel & Shmuel I. Becher, *Contracts in the Age of Smart Readers*, 90 GEO. WASH. L. REV. 83 (2022) at 118 ("[A]ll the examples used were cherrypicked. Such a selection is necessary to develop a sense of tomorrow's capabilities today. However, cherry-picking does run the risk of exaggerating the power and accuracy of the technology.") The actual capabilities of LLMs are often misrepresented; for example, during a livestream OpenAI's co-founder Greg Brockman used four examples to showcase GPT-4's capabilities, one of them involved the calculation of

even basic legal tasks.[8] It is often overlooked that despite being trained on unprecedented amounts of text, LLMs are unable to understand such text. As discussed in Part II, meaning cannot be learned exclusively from text, irrespective of its amount. One could, of course, question whether the fact that LLMs do not understand the meaning of words is of any relevance to begin with. If LLMs are capable of generating human-like, plausible sentences, it seems unimportant whether they understand text or not. After all, planes fly without mimicking birds. What is *understanding,* anyway? The attractiveness of the plane comparison fades quickly once it is realized that *generating* text is not synonymous with *using* language and that, as described in Part II, the use of language requires not only an understanding of words but an understanding of the world. Moreover, while it cannot be questioned that planes can fly (arguably, better than *some* birds!), it is questionable whether LLMs can use language.

As discussed in Part III, LLMs are trained to solve the problem of "next-word prediction."[9] They are thus not only intrinsically incapable of understanding text but also of evaluating the veracity or correctness of such text. Their resulting propensity to "hallucinate," to generate "text or responses that seem syntactically sound, fluent, and natural but are factually incorrect, nonsensical, or unfaithful to the provided source input"[10] can be regarded as a corollary to their primary language generation objective. Hallucinations can also be regarded as one of the main roadblocks to the integration of LLMs into legal workstreams. Technical limitations aside, it must be appreciated that additional roadblocks derive from the very substance of legal knowledge, namely, the difficulty of formalizing such knowledge and of establishing *a* legal ground truth against which to evaluate the generated output. In the context of legal tasks that require correctness but, at the same time, admit multiple correct answers, it may be problematic - if not altogether impossible - to determine whether the generated text constitutes a hallucination. As observed by John G. Roberts, Jr. Chief Justice of the United States:

---

taxes of hypothetical taxpayers on the basis of several sections of U.S. tax law. While Brockman proclaimed that GPT-4 can "do taxes," it later eventuated that the calculations were incorrect and that the example used was copied from an academic dataset. *see*: Libin Zhang, *Four Tax Questions for ChatGPT and Other Language Models*, 179 TAX NOTES FED. 969 (2023); Nils Holzenberger, Andrew Blair-Stanek, and Benjamin Van Durme, *A dataset for statutory reasoning in tax law entailment and question answering*, PROC. NATURAL LEGAL LANGUAGE PROCESSING WORKSHOP 31-38 (2020).

[8] Andrew Blair-Stanek, Nils Holzenberger, Benjamin Van Durne, *BLT: Can Large Language Models Handle Basic Legal Text?* ARXIV (Nov. 16, 2023) https://arxiv.org/pdf/2311.09693.pdf. (The best currently available LLMs perform very poorly at many basic legal text-handling tasks, such as looking up the text at a line of a witness deposition or at a subsection of a contract and are best compared to "very sloppy paralegals."); Matthew Dahl et al., *Large Legal Fictions: Profiling Legal Hallucinations in Large Language Models,* ARXIV (Jan. 4, 2024) https://arxiv.org/pdf/2401.01301.pdf; Sayash Kapoor, Peter Henderson, Arvind Narayanan, *Promises and pitfalls of artificial intelligence for legal applications*, ARXIV (Jan. 10, 2024) https://arxiv.org/pdf/2402.01656.pdf.

[9] *See* R. Thomas McCoy, Yao Shunyu Yao, Dan Friedman, Matthew Hardy and Thomas L. Griffiths, *Embers of Autoregression: Understanding Large Language Models Through the Problem They are Trained to Solve*, ARXIV (Sep. 24, 2023), https://arxiv.org/pdf/2309.13638.pdf.

[10] Joshua Mayne*x* et al*., On Faithfulness and Factuality in Abstractive Summarization*, PROC. 58TH ANN MEETING ASSOC'N COMPUTATIONAL LINGUISTICS 1906 (2020).

"Many professional tennis tournaments, including the US Open, have replaced line judges with optical technology to determine whether 130 mile per hour serves are in or out. These decisions involve precision to the millimeter. And there is no discretion; the ball either did or did not hit the line. By contrast, legal determinations often involve gray areas that still require application of human judgment."[11]

The law is not black and white. It is often difficult, if not impossible, to evaluate approaches to legal problems and answers to legal questions as either correct or incorrect, good, or bad. The law abounds in grey areas; cases and statutes are susceptible of multiple interpretations; many legal problems can be approached from multiple angles and many legal questions are susceptible to more than one answer. It is these gray areas are the source of most challenges – not just when determining the correctness of the generated output but, at an earlier stage, when training (or: *teaching*?) the model what the output should be.

This article does not question the value of machine-learning approaches in general. Machine learning has found many useful applications in legal practice. Examples are text analytics, legal research, predictive coding, and legal judgment prediction, to name a few.[12] In principle, however, machine learning is of limited use in the context of those tasks that require understanding and reasoning, not to mention the manipulation of substantive legal knowledge. Similarly, this article does not deny the unprecedented progress in natural language processing ("NLP") or downplay the capabilities of LLMs in *general*. Instead, it calls for a more factual approach to their capabilities. Drawing from recent findings in technical scholarship, it addresses the overly optimistic predictions as to the role of LLMs in legal practice. Integrating LLMs into legal workstreams without a better comprehension of their intrinsic limitations is bound to create inefficiencies, if not outright risks. After all, "misunderstanding how a technology works can make the difference between it being safe and dangerous."[13] Those who seek to rely on the output generated by language models should verify every sentence. It is one thing to have ChatGPT write creative copy for sales and marketing purposes; it is yet another to deploy LLMs in legal practice, where even small inaccuracies can snowball into lawsuits or result financial losses.

---

[11] John G. Roberts, Jr. Chief Justice of the United States, *2023 Year-End Report on the Federal Judiciary* (31 December, 2023) https://www.supremecourt.gov/publicinfo/year-end/2023year-endreport.pdf.

[12] Daniel M. Katz et al., *Natural language processing in the legal domain*, ARXIV (Feb. 23, 2023) https://arxiv.org/pdf/2302.12039.pdf; Serena Villata et al., *Thirty Years of Artificial Intelligence and Law: the Third Decade*, 30 AI & L. 561 (2022); Haoxi Zhong et al., *How Does NLP Benefit Legal System: A Summary of Legal Artificial Intelligence*, PROC. 58TH ANN. MEETING ASS'N COMPUTATIONAL LINGUISTICS 5218, 5222–26 (2020); Frank Pasquale & Glyn Cashwell, *Four Futures of Legal Automation* (2015) 63 U.C.L.A. L. Rev 26; Benjamin Alarie, Anthony Niblett & Albert Yoon, *How Artificial Intelligence Will Affect the Practice of Law*, 68 U. TORONTO L.J. 106 (2018); Benjamin Alarie, *The Path of the Law: Towards Legal Singularity* (2016) 66 U. TORONTO L.J. 443; For general a general overview of the application of machine learning in law, *see*: Harry Surden, *Machine Learning and Law*, 89 WASH. L. REV. 87 (2014).

[13] Oliver Brown, *Hallucinating AIs Sound Creative, but Let's Not Celebrate Being Wrong,* MIT PRESS READER (Oct. 13, 2023) https://thereader.mitpress.edu/hallucinating-ais-sound-creative-but-lets-not-celebrate-being-wrong/ .

The reader's view on the use of LLMs in the performance of legal tasks depends on which sources he or she has been exposed to. Was it the paper "Sparks of AGI,"[14] which focused on the emergent capabilities of LLMs and portrayed them as a significant step towards artificial general intelligence, or was it "Embers of Autoregression,"[15] which emphasized the challenges of adapting models trained to generate text based on stochastic word prediction to more complex downstream tasks, including those that involve reasoning, understanding and domain-specific knowledge? While the first paper made headlines, the second paper passed under the radar of the mainstream press. Yet, it is the second paper that is more relevant to practicing lawyers given that virtually all legal tasks can be regarded as downstream tasks that diverge from the  LLM's original word generation objective…

This article is organized as follows. Part I presents a brief technological background that positions LLMs within the broader field of machine learning. It focuses on the fact that machine learning models, including LLMs, learn from examples – and that the quality of the model largely depends on the quantity as well as the  quality of such examples.

Part II focuses on the main limitations of LLMs: their inability to understand text, to know and to reason. These limitations derive from the fact that LLMs are trained on large text corpora but have no access to anything outside of such corpora. It demonstrates that the symbol grounding problem, the necessity to associate words with their actual physical or conceptual referents, cannot be solved what exposure to text alone. As the ability to understand and to use language is intrinsically related to knowledge and reasoning, this part poses the seemingly trivial question *"what do LLMs actually know?"* and introduces the concept of "parametric knowledge," the information memorized in the parameters of LLMs that directly affects the quality of their output. This part also demonstrates that the low quality and incompleteness of parametric knowledge not only precludes LLMs from being able to reason but also contributes to hallucinations. These limitations are particularly important to lawyers who often perceive LLMs as knowledge bases or search engines.

Part III focuses on the quality of the generated output, the core reason why lawyers should beware of LLMs and refrain from blindly trusting their output. While LLMs can generate fluent text, such text is often factually incorrect or at least of questionable quality.[16] This part investigates the problem of hallucinations and, more specifically, the difficulty of determining *whether* the generated output constitutes a hallucination or not. It presents two contrasting scenarios, one where models are asked factual questions that yield only one possible answer and one where the question is open ended and cannot be answered with reference to a single "legal ground truth." In the second scenario, the existence of hallucinations generally depends on the expertise and opinion of the users. While it is unclear whether the problem lies in the type of question ('what should the answer be evaluated against?') or in the type of user, ('does he or she have the expertise to evaluate the answer?') it is clear that the less competent the user, the less hallucinations he or she will detect. With the exception of outputs constituting blatant nonsense, the quality – and *correctness* - of the output generated by LLMs lies in the eye of the beholder. The challenges of evaluating the quality of such output also lie in its fluency and superficial plausibility. This dangerous combination discourages verification and creates the risk of overreliance. Even experienced lawyers may have a limited (if any!) understanding of what LLMs can do and may conflate linguistic competence with domain-specific expertise.[17]

---

[14] Sebastien Bubeck et al., *Sparks of Artificial General Intelligence: Early experiments with GPT-4,* ARXIV (Apr. 13, 2023), https://arxiv.org/pdf/2303.12712.pdf. ("[W]e report on evidence that a new LLM developed by OpenAI, which is an early and non-multimodal version of GPT-4, exhibits many traits of intelligence.")

[15] McCoy et al., *supra* note 9 ("We argue that in order to develop a holistic understanding of these systems we need to consider the problem that they were trained to solve: next-word prediction over Internet text.")

Part IV aims to preclude optimistic predictions that the problem of hallucinations "will surely be solved" in the near future. It investigates the reasons why models hallucinate and the difficulty of eliminating this phenomenon. Many of the techniques seeking to adapt language models to domain-specific tasks or to mitigate the occurrence of hallucination may indirectly contribute to their existence. This part emphasizes the challenges of fine-tuning LLMs for domain-specific tasks as well as augmenting them by means of external sources of information. While fine-tuning techniques lead back to the problem of a "legal ground truth" and illustrate the difficulty of providing examples of ideal input-output pairs, augmentation methods highlight the need for reliable sources of legal knowledge. This part also discusses the technical and legal expertise required to instruct LLMs by means of prompts. Suggestions that LLMs will democratize access to justice, overlook the fact that language models will readily generate text on the basis of false premises or absurd instructions.

---

[16] Even OpenAI acknowledges that GPT-4 tends to "make up facts, to double-down on incorrect information. […] Moreover, it often exhibits these tendencies in ways that are more convincing and believable than earlier models (e.g., due to authoritative tone or to being presented in the context of highly detailed information that is accurate)," see: OPENAI, GPT-4 TECHNICAL REPORT (2023), https://cdn.openai.com/papers/gpt-4.pdf, at 9; Hussam Alkaissi & Samy I McFarlane, *Artificial Hallucinations in ChatGPT: Implications in Scientific Writing.* CUREUS (Feb. 15, 2023) https://doi.org/10.7759/cureus.35179; Daron Acemoglu, *Get Ready for the Great AI Disappointment*, WIRED (Jan. 10, 2024) https://www.wired.com/story/get-ready-for-the-great-ai-disappointment/ ("More and more evidence will emerge that generative AI and large language models provide false information and are prone to hallucination—where an AI simply makes stuff up, and gets it wrong. Hopes of a quick fix to the hallucination problem via supervised learning, where these models are taught to stay away from questionable sources or statements, will prove optimistic at best. Because the architecture of these models is based on predicting the next word or words in a sequence, it will prove exceedingly difficult to have the predictions be anchored to known truths.")

[17] Glenn Zorpette, *Just Calm Down About GPT-4 Already and stop confusing performance with competence,* IEEE Spectrum (17 May, 2023) https://spectrum.ieee.org/gpt-4-calm-down. (In this interview, the famous roboticist Rodney Brooks, people tend to mistake performance in language generation for actual competence.)

Part V addresses some anticipated objections to the broader argument that LLMs are inherently unreliable and that their output cannot be trusted. Arguably, LLMs work most of the time, or so it seems. Irrespective of their technical limitations, they achieve excellent benchmark performance, and even pass the bar exam. This part clarifies that excellent benchmark performance is not indicative of the broader abilities of LLMs and that benchmarks are hardly representative of the skills they purport to measure. Given the widespread contamination of pre-training corpora with benchmark tasks, LLMs can leverage technical shortcuts to achieve better results. Similarly, the fact that GPT-4 has passed the bar exam *once* does not translate into its general ability to support legal work.

Part VI discusses a number of potential remedial mechanisms. Opposing the view that the limitations of LLMs are only temporary and will be solved with more training data or more parameters, this part emphasizes that while 'scaled up' LLMs will further improve at word prediction, they will not "magically" acquire the ability to understand, to know or to reason. Trapped in the text of their training corpora, LLMs will remain "word calculators." This part also addresses the problem of training data, both with regards to its availability as well as to its quality. While articles commonly conclude with a series of recommendations, this part can only reemphasize the fact that many of the limitations of LLMs must be regarded as "open research questions," a concept indicating that, at the present state-of-the-art, the problems remain unsolved. The main recommendation remains. Reader beware.. LLMs can only be deployed with extreme caution – and with a full realization of their limitations.

# I. Technical Background

To understand the fascination with LLMs, it is first necessary to distinguish machine learning from traditional AI.[18]  Also known as rule-based or good old-fashioned AI (GOFAI), traditional AI is based on programs written by humans[19] and focuses on logic, knowledge, and reasoning.[20]  Traditional AI involves explicit instructions to achieve specific goals. In contrast, machine learning involves systems trained on large amounts of data to create their own programs, or models.[21]  For example, tasked to identify cats in photographs, traditional AI would involve the creation of detailed instructions how to identify cats, whereas machine learning approaches would involve the provision of examples of cats for the model to devise its "own rules" how to identify them. The first approach requires rules defining cats, the second requires thousands of cat pictures for the model to detect common features, or patterns, that represent "catness." Machine learning relies on a variety of training techniques, including supervised learning, unsupervised learning, reinforcement learning and self-supervised learning. The latter vary in the amount of human involvement required to train a model. Deep learning, a subset of machine learning inspired by the structure of the human brain, relies on artificial neural networks composed of many interconnected nodes organized in layers. [22] While supervised or reinforcement learning concern the manner models are trained, deep learning concerns their internal architecture. At present, most applications of deep learning rely on supervised learning, which requires labeled examples demonstrating the relationship between certain inputs and outputs. Notably, supervised learning is generally suited for problems where the desired output is *known and specifiable*. An example is image recognition, where the model is fed with millions of images containing a cat, each of which is labeled "cat." Once trained, the model will recognize cats in images it has not seen before. The main disadvantage of supervised learning is the need for a huge number of examples, including millions of images without a cat labeled "no cat." In unsupervised learning, the model is trained on unlabeled data, where the output is unknown. The choice of model depends on the problem as well as the type and amount of available data.

---

[18] STUART J RUSSELL & PETER NORVIG, ARTIFICIAL INTELLIGENCE: A MODERN APPROACH 4TH ED, 45-52 (2021).

[19] MELANIE MITCHELL, ARTIFICIAL INTELLIGENCE, A GUIDE FOR THINKING HUMANS 33 (2019).

[20] Gary Marcus, *The Next Decade in AI: Four Steps Towards Robust Artificial Intelligence,* ARXIV (Feb. 14, 2020) https://arxiv.org/pdf/2002.06177.pdf.

[21] Yann LeCun, Yoshua Bengio, Geoffrey Hinton, *Deep Learning,* 521 NATURE 436 (2015).

In principle, a machine learning model is an expression of an algorithm that identifies patterns in data. [23] The model itself is the product of the training process and its quality in terms predictive capabilities largely depends on the quality of its training. In the context of LLMs, which are a particular type of machine learning model, the training data is text and the training objective is to model language or, more specifically, the probability distributions of language.[24] They do so by predicting the next word by assigning probabilities to sequences of words that are likely to occur in a given language.[25] The recent breakthroughs in LLMs are commonly associated with so-called transformer models, which enable the training of larger and deeper neural networks and with the introduction of the attention mechanism.[26] Unlike language models based on n-gram models, which can only predict words given the preceding sequence of 5 or fewer words, the attention mechanism evaluates relationships between multiple words, regardless of their respective positions, and allows language models to generate text that *appears* coherent and plausible over multiple paragraphs. In simpler terms, unlike earlier langue modelling architectures, LLMs can "ingest" and process larger amounts of text. More importantly, they can mimic human language without knowing the rules of grammar or syntax.[27]

---

[22] The word "deep" refers to the fact that the "circuits are typically organized into many layers, which means that computation paths from inputs to outputs have many steps." Russel & Norvig, *supra* note 18 at, 800, 926.

[23] Yann LeCun, Yoshua Bengio, Geoffrey Hinton, *Deep Learning,* 521 NATURE 436 (2015).

[24] Russel and Norvig define a language model "as a probability distribution describing the likelihood of any string. Such a model should say that 'I don't disturb the universe?' has a reasonable probability as a string of English, but 'universe, dare the I disturb do?' is extremely unlikely." *See:* Russel & Norvig, *supra* note 18, at 875.

[25] AJAY AGRAWAL, JOSHUA GANS & AVI GOLDFARB, PREDICTION MACHINES: THE SIMPLE ECONOMICS OF ARTIFICIAL INTELLIGENCE (2018).

[26] Ashish Vaswani et al. *Attention is All you Need*, 31 CONF. ON NEURAL INFO. PROCESSING SYS. (NIPS 2017) (Attention mechanisms enable models to focus on more relevant parts of the input and detect long-range dependencies between words.)

LLMs are trained in two stages, commonly referred to as pre-train and fine-tune.[28] LLMs are pre-trained on large corpora of *unlabelled* text, such as millions of books or online resources. In this unsupervised phase, models learn the *general* structure of language and acquire the ability to predict the next word. Although pre-training requires an immense amount of data and computational resources, it does not require such data to be labelled. [29] Next, models are fine-tuned on smaller, domain- or task-specific datasets. Fine-tuning requires fewer computational resources than pre-training and a significantly smaller amount of training data. It does, however, require more human input in the preparation of datasets.

Once trained, LLMs can be instructed  by their users by means of so-called "prompts," strings of natural language text that guide models to perform various downstream tasks, such as translation, summarization, sentiment analysis or question answering, to name a few.[30] In fact, the common fascination with transformer-based LLMs can be attributed to the fact that although they are trained to predict the next word, they do surprisingly well at a wide range of other language tasks.[31]

Three points are worth remembering throughout the discussion:

First, while LLMs are data-driven and purely statistical in nature, it must not be assumed that most problems can be solved with providing them with more data or that the quality of such data can be disregarded.[32] In its seminal book, *The Master*

---

[27] Russel & Norvig, *supra* note 18, at 928.

[28] Russel & Norvig, *supra* note 18, at 832.

[29] Alec Radford et al., *Improving Language Understanding by Generative Pre-Training* (2018) https://cdn.openai.com/researchcovers/languageunsupervised/language_understanding_paper.pdf.

[30] Tom B. Brown et al., *Language Models Are Few-Shot Learners*, 34 CONF. ON NEURAL INFO. PROCESSING SYS. (2020); Laria Reynolds and Kyle McDonell, *Prompt Programming for Large LLMs: Beyond the few-shot paradigm*, PROC. 2021 CHI CONF. HUMAN FACTORS COMPUTING SYSTEMS  (2021); Timo Schick and Hinrich Schütze, *Exploiting cloze-questions for few-shot text classification and natural language inference*, PROC. 16TH CONF. EU. CH. ASS'N COMPUTATIONAL LINGUISTICS 255–269 (2021).

[31] Typical natural language tasks include speech recognition, machine, translation, information extraction  (i.e., the process of acquiring knowledge by skimming a text and looking for occurrences of particular classes of objects, and for relationships among them), information retrieval (i.e., the task of finding relevant documents or sources, as exemplified by search engines. Question answering, however, must be regarded as a different task altogether given that responses do not take the form ranked lists of documents but require actual answers, *see:* Russel & Norvig, *supra* note 16, at 900, 901.

[32] Russel and Norvig describe the challenges of obtaining enough data to build a robust model and preparing such data to train the model. They observe that in natural language processing, it is more common to work with unlabeled text due to the difficulty of labeling: an unskilled worker can easily label an image as a "cat" or "motorbike," but requires extensive training – if not a legal degree -  to annotate contractual clauses as, for example, compliant with local consumer protection regulations. *supra* note 16, at 922; *see also*: Brown et al., who states that a major limitation to the pre-

*Algorithm*, Professor Domingos observed that "if machine learning was something you bought at the supermarket, its carton would say: 'just add data.'"[33] He also stated, however, that "no matter how good the learning algorithm is, it's only as good as the data it gets."[34] As described below,[35] many of the shortcomings of LLMs derive from the poor quality of the data they are pre-trained on and fine-tuned with.

Second, although LLMs can be adapted to a range of downstream tasks that diverge from their original training objective, there may be a "mismatch between the problem that a system developed to solve and the task that it is being given."[36] In principle, a system "confined to token-level, left-to-right" word prediction[37] may not always seamlessly perform other tasks, especially if such tasks require domain-specific knowledge or reasoning.

Third, different LLMs have varying capabilities and are better suited for different tasks.[38] For example, autoregressive decoders, such as GPT-4, excel at text generation and some types of summarization. Their learning is, however, unidirectional as the model can only use information from the left of the predicted token. Autoregressive decoders are thus inherently limited to predicting the next word and, *by definition*, cannot understand the broader context of the generated text. Bidirectional auto-encoders, such as BERT, predict masked words from a sentence based on context from both before and after the missing word.[39] Theoretically, they are more suitable for tasks that require an understanding of larger inputs, such as sentence classification and information extraction. It is worth observing that until the arrival of ChatGPT and such LLMs like GPT-4, Claude 3 or Llama 2, most research in legal natural language processing ("NLP") focused on BERT.[40]

---

train and fine-tune approach is that "while the architecture is task-agnostic, there is still a need for task-specific datasets and task-specific fine-tuning: to achieve strong performance on a desired task typically requires fine-tuning on a dataset of thousands to hundreds of thousands of examples specific to that task. *supra* note 30, at 3, 8-9.

[33] PEDRO DOMINGOS, THE MASTER ALGORITHM 7 (2015).

[34] *See id.* at 45.

[35] *See infra* Section IV.B.

[36] McCoy et al., *supra* note 8, at 4-7 ("The tasks for which LLMs are used often differ from the problem they were trained to solve. For example, even though they were trained for next-word prediction, LLMs are sometimes asked to translate sentences. Why does this matter? When a system is adapted for one purpose but then co-opted for a different purpose, the original purpose may influence the system's nature in ways that would not make sense if only the new purpose were considered.")

[37] Xiaofei Sun et al., *Pushing the Limits of ChatGPT on NLP Tasks,* ARXIV (Oct. 9, 2023) https://arxiv.org/pdf/2306.09719.pdf (describing the mismatch between ChatGPT and many NLP tasks, which cannot be easily formatted as a text generation task. The adaptation of the original text generation task to other tasks often comes at a heavy cost in performance).

[38] DAVID FOSTER, GENERATIVE DEEP LEARNING, 2ND ED. (2023)

[39] Autoregressive models, such as GPT-3, process text from left to right and assign probabilities based only on the preceding text. In contrast, bidirectional models learn from the surrounding text on both sides of the target word. *See:* Jacob Devlin, Ming-Wei Chang, Kenton Lee & Kristina Toutanova, *BERT: Pre-training of Deep Bidirectional Transformers for Language Understanding,* PROC. 2019 ANN. CONF. N. AM. CH. ASS'N COMPUTATIONAL LINGUISTICS 4171 (2019).

[40] Lucia Zheng, Neel Guha, Brandon R. Anderson, Peter Henderson & Daniel E. Ho, *When Does Pretraining Help? Assessing Self-Supervised Learning for Law and the CaseHOLD Dataset of 53,000+ Legal*

## II. Language Models Do Not Understand

The fact that LLMs can emulate human language is fascinating.[41] Nonetheless, the ability to generate plausible text must be distinguished from the ability to *understand* and to *use* language. This distinction is particularly important in the legal field as the performance of most legal tasks requires a mastery of language, not just the generation of text.

### A. Understanding Language

In principle, a language model is "a system for haphazardly stitching together sequences of linguistic forms it has observed in its vast training data, according to probabilistic information about how they combine, but without any reference to meaning: a stochastic parrot."[42]  It must be appreciated that despite unprecedented progress in natural language *generation*, natural language *understanding* remains one of the grand challenges in AI. This point is frequently made in technical scholarship[43] but is rarely, if ever, acknowledged in legal scholarship.[44] At a basic level, LLMs understand neither their input nor their output. The do not know the meaning of words. This limitation derives from their training objective and from their training method. As indicated, the primary objective of an LLM is statistical word prediction and the main method of training LLMs to achieve this objective is to expose them to enormous amounts of text. During pre-training, models learn to predict the next token based on the previous token(s) or, more specifically, to produce a "reasonable continuation" of the preceding text by estimating the probabilities of the next word, based on prior words, including those generated by itself.[45]

---

More importantly, the inability to understand text seems to derive from the fact that LLMs are trained on text and have no access to anything outside of such text. In a seminal 2022 article on natural language understanding, Bender and Koller claim that LLMs cannot learn the meaning of words without access to the world that exists outside of their training corpus.[46] Meaning cannot, as they claim, be learned from form alone. The authors define "form" as any observable realization of language, such as text, and "meaning" as the relation between the form and something external to language.[47] LLMs cannot learn the meaning of words, or text, if they cannot relate such words to something that exists outside of the training corpus, something other than the text itself. The size of the training corpus is irrelevant in this regard. Understanding requires grounding: a connection between the text and the physical reality, an association between words and the world. The meaning of words on a page is said to be "ungrounded," whereas the meaning of words in one's head — words one already *understands*— is "grounded." For LLMs to understand words, it would be necessary to solve the perennial symbol grounding problem,[48] that is, to train them to associate symbols, such as words, with their real-world referents, such as physical objects.

Inundated with headlines about the progress of LLMs and with incessant predictions "how Generative AI will revolutionize work,"[49] it is easy to forget that the symbol grounding problem remains a fundamental challenge in NLP given that the symbolic representations of objects and concepts, the *words* representing things and concepts, are always arbitrary.[50] The word "apple," for example, is a string of letters with no inherent meaning. The meaning of the word "apple" derives solely from the association of this string with the object commonly recognized as an apple. Persons who had never seen an apple may have trouble associating the word with the object – irrespective of the number of times they have seen the word "apple." Unless one holds the fruit in front of their eyes or points at it at the supermarket, the question "would you like an apple?" will be incomprehensible. To an LLM, such question will be nothing but an abstract string of words.

for a more technical elaboration of the training objectives of transformer-based LLMs, *see*: McCoy et al., *supra* note 8, at 5-7.

[46] Emily M. Bender & Alexander Koller, *Climbing towards NLU: On Meaning, Form, and Understanding in the Age of Data*, PROC. 58TH ANN. MEETING ASS'N COMPUTATIONAL LINGUISTICS 5185 (2020).

[47] Bender & Koller, *supra* note 46, at 1 ; LUDWIG WITTGENSTEIN, PHILOSOPHICAL INVESTIGATIONS § 43 (1953) (who associated the meaning of a word with its use in language).

[48] Stevan Harnad, *The Symbol Grounding Problem*, 42 PHYSICA D: NONLINEAR PHENOMENA, 335-346 (1990), Selmer Bringsjord, *The Symbol Grounding Problem… Remains Unresolved* 27 J EXPERIMENTAL & THEORETICAL ARTIFICIAL INTELLIGENCE 63-72 (2015); Grounding can also be described as "the connection between logical reasoning processes and the real environment in which the agent exists," *see*: Russel & Norvig, *supra* note 16, at 234.

[49] *Generative AI could radically alter the practice of law*, THE ECONOMIST, (June 6, 2023) https://www.economist.com/business/2023/06/06/generative-ai-could-radically-alter-the-practice-of-law.

[50] Mapping from symbols to objects is not formally defined – in natural language, two occurrences of the same word or phrase may refer to different things in the world, see: Russel & Norvig, *supra* note 18, at 875.

The problem of symbol grounding gains in complexity in the case of abstract referents, like "justice" or "indemnity." In such instance, meaning does not derive from associations with physical objects but from relationships between concepts.[51] Even then, however, such relationships require internal representations that exist independently from their textual representations and, at some stage, require grounding in the physical world. An LLM could, for example, ingest millions of sentences containing the word "indemnity," including thousands of indemnity clauses. An understanding of the word "indemnity" would, however, require prior familiarity with such concepts like "liability," "obligation," "risk," or "loss." While an LLM could, again, be provided with millions of sentences containing these words, at some stage, it would have to understand the more fundamental concept of causation: it would have to understand that actions or events in the physical world have consequences, including damage to objects or injuries to people. It would also have to understand that while many events remain beyond human control, most actions are attributable to people and that in most legal systems people are held accountable for their actions.

Theoretically, the grounding problem could be addressed by training LLMs on text augmented with visual information about the world[52] or by exposing them to virtual worlds.[53] Such approaches are, however, extremely difficult to implement in practice given the resources required to enrich text with images or videos representing common objects as well as the relationships between such objects or to create virtual environments conveying the complexity of the real world. After all, even a simple contract of sale requires an understanding of the various goods that can be the object of sale not to mention their specific attributes. Is the object easily damaged? Will it decay without refrigeration? Does it require special packaging? In effect, the images and videos accompanying the text would have to convey the characteristics of such objects – from their shape, weight, and size, down to such details like durability or viscosity. Moreover, it would also be necessary to provide representations of spatial relationships between the goods and the objects that commonly serve to transport, contain, or transfer such goods: boxes, warehouses, or counters. Without an understanding of such relationships LLMs would not be able to master the concepts of "delivery" or "transfer of risk." Is it, however, feasible to create a virtual world representing such objects, not to mention all the possible configurations between them? What about more complex transacting scenarios, such as international contracts of sale that involve loading docks, shipping containers and sea routes threatened by pirate attacks? How many embedded images or virtual worlds would it take for models learn the concepts of "possession," "transfer" or "distance"?

---

[51] Steven T. Pintadosi & Felix Hill, *Meaning without Reference in Large Language Models,* ARXIV (Aug. 12, 2022) https://arxiv.org/pdf/2208.02957.pdf.

[52] Yonatan Bisk et al., *Experience Grounds Language*, PROC. 2020 CONF. EMPIRICAL METHODS IN NLP 8718 (2020) (physical and social context is indispensable for language models to genuinely understand language).

[53] Thomas Carta et al, *Grounding Large Language Models in Interactive Environments with Online Reinforcement Learning*, PROC. 40TH INT'L CONF MACHINE LEARNING 3676 (2023).

Bender & Koller emphasize that irrespective of the amount of text and the length of training, language models trained exclusively on text cannot develop the ability to associate words with objects or, more generally, learn about the world.[54] Without solving the symbol grounding problem, LLMs will not evolve beyond stochastic pattern recognition and will remain perpetually confined to a world of text. This observation leads to the next point: the ability to use language.

## B. Using Language

It is useful to distinguish formal linguistic competence, which denotes knowledge of the statistical regularities of language, from functional linguistic competence, which denotes knowledge of how to use language in real life situations.[55] Although formal linguistic competence could be regarded as a prerequisite of functional competence, the latter seems more important than the former as it leads to the accomplishment of specific goals in the real world. Humans generally do not generate text for the sake of it but use language as a tool to communicate, to socialize, to solve problems, to express feelings etc. This point is particularly important in the legal profession. It is true that lawyers consume and produce a lot of text. Nonetheless, leaving aside the production of lengthy documents to extract higher hourly fees or to prolong expensive litigation, lawyers do not generate text but use language for purpose-driven communications that serve specific legal tasks. Text serves as a medium to provide legal advice, to win arguments, to allocate risks or to fulfill the procedural demands of litigation, amongst others. In the legal context, text must therefore be seen as a means of conveying knowledge and expertise – not as an end in itself. In contrast, while LLMs excel at formal linguistic competence, they are unable to use language to achieve certain objectives or to solve problems in real-life scenarios.[56] Their main objective being word prediction, LLMs do not have goals of their own and cannot understand the goals of others.

The effective use of language requires "functional grounding," the ability to predict and control physical as well as social processes.[57] The concept is particularly useful in the present discussion as it ties the *use* of language to the broader concepts of knowledge and reasoning.[58] It is difficult to imagine, after all, how one could use

---

[54] Bender & Koller, *supra* note 46, at 9; Margaret Mitchell, *What Does It Mean for AI to Understand?* QUANTA (Dec. 16, 2021) https://www.quantamagazine.org/what-does-it-mean-for-ai-to-understand-20211216/ (Professor Mitchell observes that "understanding language requires understanding the world, and a machine exposed only to language cannot gain such an understanding." "Consider what it means to understand 'The sports car passed the mail truck because it was going slower.' You need to know what sports cars and mail trucks are, that cars can 'pass' one another and, at an even more basic level, that vehicles are objects that exist and interact in the world, driven by humans with their own agendas.")

[55] Kyle Mahowald et al., *Dissociating Language and Thought In Large Language Models: a Cognitive Perspective,* ARXIV (Nov. 4, 2023) https://arxiv.org/pdf/2301.06627.pdf

[56] For discussion of this point see: Pintadosi & Hill, *supra* note 51.

[57] Thomas Carta et al, *Grounding Large Language Models in Interactive Environments with Online Reinforcement Learning,* PROC. 40TH INT'L CONF. MACHINE LEARNING 3676 (2023).

[58] Mahowald et al., *supra* note 55, at 14.

language without basic world knowledge, including knowledge of objects and their attributes. It is equally difficult to imagine how language could be used without the ability to reason – if only about such simple facts like "flowers require refrigeration to be transported over longer distances" or "if Charlie loses his job, he may not be able to pay off his car loan." As in the case of symbol grounding, functional grounding requires a physical experience of the real world that would lead to the development an internal world model. Needless to say, the relationships between understanding, knowledge and reasoning are complex and susceptible of endless philosophical explorations. From a practical perspective, however, the three concepts are difficult to disassociate and can be regarded as set of minimal prerequisites for the effective use of language. After all, in many instances the verb "to understand" is easily replaced with the verb "to know." Humans *understand* the word "apple" because they *know* what an apple is…

As the symbol grounding problem cannot be solved by training models on more text and as it is widely accepted that LLMs lack world knowledge,[59] the discussion could end here. It is, after all, reasonable to assume that legal knowledge, a more formalized and explicit type of knowledge, cannot be operationalized without world knowledge - or basic common sense. How could a system that does not understand language and lacks common sense augment legal work, not to mention – replace lawyers? How could it analyze contracts, suggest litigation strategies, or answer legal questions pertaining to concrete cases? All these tasks require not only legal knowledge but also a solid grasp of the physical world (and - common sense)! Nonetheless, given that technical literature often discusses knowledge, common sense, and reasoning independently, it is necessary to briefly address each of the aforementioned skills separately – if only to better illustrate the inherent limitations of LLMs.

## C. Knowledge

Although LLMs are neural networks trained to model word distributions, they are often perceived and consequently used "as if" they were knowledge repositories, databases, or even search engines. As in many instances LLMs *seem* to generate adequate output that satisfies the user's objective,[60] it is necessary to ask: what do LLMs actually know? Can they know anything? Is it possible to speak of knowledge in the first place? It is worth re-emphasizing that, in principle, LLMs are trained to extract statistical information about word distributions and are best described as "word calculators,"[61] not knowledge bases. Nonetheless, it must be acknowledged that given the sheer size of their training corpora, LLMs memorize strings of co-occurring words contained in such corpora.[62]  Moreover, as LLMs often produce meaningful as well as correct answers to user queries[63] and must be deemed to "learn" or "assimilate" *some* knowledge,  it is worth introducing  the concept of parametric knowledge, the knowledge encoded in the parameters of an LLM.[64]

---

[59] Talmor et al., *CommonsenseQA 2.0: Exposing the Limits of AI through Gamification*, NeurIPS (2021); this difficulty pertains to computers in general, not just to language models, see: JOHN HAUGELAND, ARTIFICIAL INTELLIGENCE, THE VERY IDEA 176 (1989).

[60] For an explanation why language models may only seem to fulfill the user's objective, *see infra* Section III.A.

[61] S. Willison, *Think of language models like chatgpt as a "calculator for words"* (Apr 2023), https://simonwillison.net/2023/Apr/2/calculator-for-words/

In technical terms, parametric knowledge derives from positionally close, or highly co-occurring word associations explicitly contained in the text corpora used in training.[65]  If certain word strings occur more often than others, then the neural connections associated with such strings are strengthened and subsequently stored in the model's parameters.  Consequently, parametric knowledge is not encoded as explicit facts or rules but as weights representing connections in neural networks.[66] In practice , if a particular word pattern is popular in the training corpus, it will be memorized by the model and is likely to appear in – or at least affect - the generated output.[67] As LLMs have billions of parameters, they can capture a wide variety of relationships between words, phrases, and even broader language structures. These words, phrases and language structures may, in turn, reflect many facts or opinions that the model has encountered in the training data. It bears emphasizing that the retention, or memorization, of such facts or opinions in the model's parameters depends exclusively on their popularity, the number of times the sequences of words representing such facts or opinions occur in the training corpus, not on their correctness or veracity.[68] LLMs operate exclusively at the level of word distributions, not at the level of substantive facts or verified knowledge. LLMs are sensitive to the probability of words,[69] not to the veracity or correctness of statements expressed by means of such words.  If, for example, the word sequence "contracts are enforceable

---

[62] Bommasani et al., *supra* note 3, 49; Nicholas Carlini et al., *Extracting Training Data from Large Language Models*, PROC. 30TH USENIX SEC. SYMPOSIUM 2633 (2021) (language models can memorize specific examples found in their training data); Zhengbao Jiang, Frank F. Xu, Jun Araki & Graham Neubig, *How Can We Know What Language Models Know?*, 8 TRANSACTIONS ASS'N COMPUTATIONAL LINGUISTICS 423, 423–25 (2020) (overview of the challenges of examining the knowledge embedded in language models).

[63] Dan Hendrycks et al., *Measuring Massive Multitask Language Understanding*, 9TH INT'L CONF. LEARNING REPRESENTATIONS at 6 (2021) MMLU is specifically designed to evaluate the knowledge acquired during pre-training of the model by including only zero-shot and few-shot learning tasks. It contains about 16K multiple-choice questions divided into 57 subtasks, covering subjects in the humanities, social sciences, hard sciences, and other areas.

[64] The term "parameter" refers to the connections between layers in a neural network; Adam Roberts et al., *How Much Knowledge Can you Pack into the Parameters of a Language Model?*, CONF. EMPIRICAL METHODS IN NATURAL LANGUAGE PROCESSING 5416 (2020) https://aclanthology.org/2020.emnlp-main.437.pdf.

[65] Yue Zhang et al. *Siren's Song in the AI Ocean: A Survey on Hallucination in Large Language Models,* ARXIV (Sep. 24, 2023) https://arxiv.org/pdf/2309.01219.pdf. at 10, 11.

[66] Fabio Petroni et al., *Language Models as Knowledge Bases?* PROC. 9th INT'L. JOINT CONF. NLP 2463 (2019).

[67] McCoy et al., *supra* note 8, at 47. ("LLMs are biased toward sentences that have a high probability. The probability of a sentence is determined by the particular dataset that the LLM was trained on. Consequently, if the dataset contains some sentences that are frequently repeated, the model is likely to memorize them—even if they are not important or high-probability sentences in the broader world.")

[68] McCoy et al., *supra* note 9, at 2.

[69] McCoy et al., *supra* note 9, at 8.

agreements" appears in the training corpus with sufficient frequency, the network will strengthen the relevant connections between neurons and assign a higher probability to the word 'agreement' if it is preceded by the words 'contract' and 'enforceable.'[70] The model will learn that these words should appear together and, when responding to the prompt "what is a contract?," it will likely generate text containing the aforementioned string of words. It will do so because this particular sequence of words has been stored in its parameters not because the LLMs *knows* the law or *understands* what contracts are. If, however, the word sequence "contracts require *mens rea*" was abundantly present in the training corpus, the model would likely generate an answer describing *mens rea* as a pre-requisite of enforceability! The fact that *mens rea* is unrelated to contract law and denotes the intention or knowledge of wrongdoing in criminal law would be irrelevant. Irrespective whether one agrees with the term parametric *knowledge*, what matters is that such "knowledge" is neither reliable nor accurate. Arguably, given its statistical "pedigree" and the fact that it pertains to a machine learning model, a better term might be parametric *memory*…

Apart from its questionable accuracy, parametric knowledge is also inherently incomplete as does not contain the full set of facts about the world, not to mention complete restatements of specialized domains, such as law. While it can contain mathematical (e.g., "two plus two is four"), factual (e.g., "birds fly") or even legal ("contracts are enforceable agreements") information, it will not contain such basic facts, or world knowledge, like "wheels are round" or "vases thrown out of the window will break." The reason for this shortcoming derives from the fact that statements conferring such information do not occur in the training corpora with sufficient frequency to be stored in the model's parameters. Arguably, many of the basic facts about the world are not mentioned in such corpora at all. Why would anyone state the obvious? Why would Wikipedia entries, US patents and news websites, three popular sources of training data,[71] be devoted to facts that are commonly known and require no elaboration? The incompleteness of parametric knowledge is not attributable to the type or the amount of training data but to the simple fact that knowledge is not a purely linguistic phenomenon. A lot of knowledge it is not expressed in and therefore cannot be learned from text; a lot of knowledge is implicit and derives from experiencing the physical world.[72] After all, humans learn to read long after they have "learned about" causality, gravity, and about the persistence of objects in space. According to Yann LeCun, one of the fathers of deep learning, "as LLMs have no direct experience with reality, the type of common-sense knowledge they exhibit is very shallow and can be disconnected from reality."[73]

---

As an aside it is worth observing that there have been multiple attempts to create knowledge bases containing the general rules of common sense as well as fundamental facts about the world.[74] The famous CYC project, for example, which has aimed to painstakingly formalize common world knowledge, has occupied a team of knowledge engineers since 1984. CYC contains rules stated in a sufficiently general and domain-independent form to provide a universal "reasoning backbone" for other AI systems.[75] Arguably, the success of integrating LLMs with external knowledge bases would largely depend on the completeness and accuracy of such knowledge bases. At present, such integration remains an open research question[76] and provides a constant reminder that without access to external sources of knowledge, not to mention access to the physical world, LLMs cannot evolve from systems that generate text to systems that use language…

## D.  Reasoning

It is difficult to imagine the ability to use language without the ability to reason – and it is generally accepted that tasks that require reasoning still remain beyond the reach of deep-learning models, "no matter how much data you throw at them."[77] Reasoning cannot be learned from text alone as the rules of reasoning cannot be written down[78] or replicated by feeding language models with formal logic.[79] More importantly, it is impossible to reason without a "rich background knowledge about how the physical and social world works"[80] or, as observed by LeCun, a collection of world models informing the LLM "what is likely, what is plausible, and what is impossible."[81] One cannot reason about something one does not know or to make use of knowledge without the ability to reason.[82] The impossibility to separate reasoning from a background of world knowledge can be illustrated with some simple examples of reasoning on the basis of the inference framework. The inference framework, which is often regarded as a template for basic formal reasoning, comprises the processes of deduction, induction, and abduction.[83] Deduction entails a general rule which is applied to a concrete case, whereas induction involves learning and forming generalizations from a multitude of examples. Abduction involves the selection of the best explanatory hypotheses for partial observations.[84]

---

[74] Other attempts at formalizing common sense and world knowledge include Mosaic and Comet; for an overview of such projects, *see:* https://mosaic.allenai.org.

[75] The commonsense knowledge in the Cyc KB spans every domain of human experience, captures and can reason with the fundamental rules of thumb about "how the world works," such as no two objects can occupy the same space simultaneously; *see*: Douglas B. Lenat et al., *CYC: Towards Programs with Common Sense*, 33 COMM. ACM. 30 (1990).

[76] For a recent example see: Zhou Wangchunshu, Ronan Le Bras & Yejin Choi, *Commonsense Knowledge Transfer for Pre-trained Language Models,* ARXIV (Jun. 4, 2023) https://arxiv.org/pdf/2306.02388.pdf (observing that language models demonstrated limited capabilities in acquiring implicit common-sense knowledge from pre-training alone, compared to learning linguistic and factual knowledge that appear more explicitly in the surface patterns in text.)

[77] Francois Chollet, *Deep Learning with Python* (2018 Manning) 325; Alessio Plebe and Giorgio Mario Grasso, *The Unbearable Shallow Understanding of Deep Learning*, 29 MINDS & MACH. 515 (2019); Marco Lippi, *Reasoning with Deep Learning: an Open Challenge,* URANIA@AI*IA (2016); Fei Yu, Hongbo Zhang, Benyou Wang, *Natural Language Reasoning, A Survey,* ARXIV (May. 13, 2023) https://arxiv.org/pdf/2303.14725.pdf; Matthew Hudson, *The Language Machines* (2021) 591 NATURE 22.

[78] MICHAEL POLANYI, PERSONAL KNOWLEDGE: TOWARDS A POST-CRITICAL PHILOSOPHY, Ch. 5 (1958).

[79] Yejin Choi, *The Curious Case of Commonsense Intelligence,* 151 DAEDALUS 139, 144 (2022).

[80] *Id* at 139, 142, 143.

[81] Yann Lecun, *A Path Towards Autonomous Machine Intelligence* (27 June 2022) https://openreview.net/forum?id=BZ5a1r-kVsf  3; Tyler Millhouse, Melanie Moses, Melanie Mitchell, *Embodied, Situated, and Grounded Intelligence: Implications for AI,* ARXIV (Oct. 24, 2022) https://arxiv.org/pdf/2210.13589.pdf.

It is surprisingly easy to demonstrate that without world knowledge, LLMs cannot handle even simple cases of deduction. A representative case of deduction, known as *modus ponens*, contains two statements known to be true, which lead to a conclusion.[85] The two statements comprise a major premise, the rule, and a minor premise, the facts. If, for example, the rule is that "whenever it rains, the streets are wet" and if it is raining, then it can be deduced that the streets are wet. If it is not raining, the argument does not apply but remains valid. If it is raining, the argument becomes sound. Validity denotes formal correctness; soundness preserves the truth.[86] Needless to say, lawyers require more than just formal correctness – they require the truth!  Deduction provides a blueprint for correct reasoning and preserves the truth - but only if the premises are true. The problem is that sound arguments require true premises and that LLMs cannot know and are incapable of verifying whether the premises are true. They can memorize a rule but are fundamentally unable to determine whether such rule makes sense in the first place. Provided with the rule "If it rains, then pigs fly" and with the fact that "it is raining," an LLM may "deduce" that the sky is full of pigs! LLMs do not *know* that pigs do not fly. It is unlikely that any training corpus contained an exhaustive list of non-flying animals or that a model's parametric knowledge "states" that pigs do not fly. The fact that LLMs lack basic world knowledge,[87] presents particular challenges when they are used for legal tasks that require true premises.[88]

Regarding induction, it is true that *by definition* LLMs excel at automated induction as they learn to predict the next word based on billions of examples of word patterns. Unfortunately, LLMs often fail to generalize from such examples and "overfit" on the training data, that is: they memorize patterns instead of learning broader rules. Arguably, the concept of parametric knowledge can be regarded as an example of such memorization. It also bears emphasizing that knowledge learned from examples is always provisional and depends on their number as well as on their quality. Such knowledge is also, by definition, confined to what the model has seen before.  It follows that LLMs are incapable of dealing with novelty, at least not in the sense of being able to extrapolate from prior experiences and to apply a general rule to a new situation. This shortcoming alone could be regarded as a roadblock to a wider reliance on data-driven approaches, including those exemplified by language models, in the legal domain. A language model that cannot deal with novel scenarios and can only replicate what has been done before seems of rather limited use. An even more significant roadblock, however, lies in the fact that induction only reveals correlations and cannot lead to models inferring and *thus learning* causal relationships from such correlations.[89] According to Professor Judea Pearl, who is often credited with developing a theory of causal and counterfactual inference, it is impossible to learn causal inference from observing data alone.[90] The problem is that causal inference requires a certain "amount" of pre-existing world knowledge – knowledge that cannot be acquired from examples. After all, induction is based only on enumeration and does not require knowledge or an explanation *why*, to use a popular example, all men are mortal! As induction is associated with learning from experience[91] then it must be remembered that all LLMs ever experience is text. These limitations of induction are inherent and cannot be improved with more examples.

---

[82] ERIK J. LARSON, THE MYTH OF ARTIFICIAL INTELLIGENCE: WHY COMPUTERS CAN'T THINK OF THE WAY WE DO 175 (2021).

[83] For an overview of how AI fails at inference *see*: Larson, *supra* note 82, at 104-120.

[84] Choi, *supra* note 79, at 145.

[85] PETER SMITH, AN INTRODUCTION TO FORMAL LOGIC 2, 3  (2003).

[86] *Id* at 14.

[87] An archive of LLM failures can be found at https://github.com/giuven95/chatgpt-failures.

[88] *See infra* Section IV.D.

Lastly, abduction provides a means of explaining how people choose one hypothesis from an indefinite number of possible hypotheses[92] or, in simpler terms, draw the simplest and most likely conclusion from a set of examples. Unlike induction, which moves from examples to generalizations that provide uncertain knowledge, abduction moves from observations of a particular fact to a hypothesis that explains it – sometimes with a considerable amount of guesswork and creativity![93] Translated into a legal scenario, when encountering a case that could be approached with a number of different precedents, judges will abduce a rule that "sorts and explains" the individual cases. After being assessed against a set of external criteria, the rule is either confirmed or rejected. Consequently, as in the case of deduction and induction, abduction requires reliance on some pre-existing or contextual knowledge that enables the evaluation of the individual hypotheses.

Most of the above arguments could be made with regards to reasoning by analogy, which is particularly relevant in the legal context and involves a "non-identical or non-literal similarity comparison between two things, which has a predictive or explanatory effect."[94] In legal practice, analogical reasoning involves the detection of the similarities between two or more cases and their subsequent adaptation to new cases.[95] Logically, the detection of those similarities that are relevant to a given case requires not only a good grasp of legal principles but, first and foremost, a fair amount of world knowledge. How could an LLM discover the *relevant* similarities or compare the facts of multiple cases if such facts are described by means of words that the model does not understand?

---

[89] Zhijing Jin et al., *Can Large Language Models Infer Causation from Correlation?* ARXIV (Dec. 31, 2023) https://arxiv.org/pdf/2306.05836.pdf.

[90] JUDEA PEARL, THE BOOK OF WHY 36 (2018); Domingos observes there is no such thing as learning without knowledge. "Data alone is not enough. Starting from scratch will only get you to scratch. Machine learning is a kind of knowledge pump: we can use it to extract a lot of knowledge from data, but first we have to prime the pump." Domingos, *supra* note 33, at 64.

[91] Larson, *supra* note 82, at 115.

[92] Dan Hunter, *Reason is Too large: Analogy and Precedent in Law*, 50 EMORYLJ 1197, 1259, 1260 (2001).

[93] Larson describes how the grandfather of abduction, Charles Sanders Pierce, found the thief of his stolen valuables on the basis of an intricate plan that confirmed his prior guesswork. In other words, before devising a plan, he had a "hunch" as to who the thief was! *See*: Larson, *supra* note 82, at 159-161; *see also*: Choi, *supra* note 79, 145 ("What is remarkable about abductive reasoning is that it is a form of creative reasoning: it generates new information that goes beyond what is provided by the premise. Thus, abductive reasoning builds on our imaginative thinking, which, in turn, builds on our rich background knowledge about how the world works.")

A terminological remark is apposite. Many technical papers as well as the popular press often state that language models understand[96] or reason.[97] Such statements seemingly contradict the broader academic consensus that LLMs are incapable of either.[98] It must be remembered that the terms "understand" and "reason" are often used colloquially, particularly in promotional materials[99] or narrowly, in technical contexts[100] that equate these terms with the improved performance of narrowly defined tasks such as textual entailment or document classification.[101] If, for example, an LLM can perform any of the aforementioned tasks above random chance, it is said to "understand" or "reason."[102] LLMs have also been evaluated on various reasoning benchmarks including common-sense, logical, and even ethical reasoning.[103] Given the simplistic nature of their constituent tasks, the fact the some LLMs can pass some of the existing benchmarks cannot support broader claims concerning their general ability to reason – especially with regards to tasks that require complex or multi-step reasoning.[104]

---

[94] Hunter, *supra* note 92, at 1206.

[95] Hunter, *supra* note 92, at 1252, 1256.

[96] For example, see multiple express references to understanding in the seminal paper by Alec Radford et al., *Improving Language Understanding by Generative Pre-Training,* (2018) ("We introduced a framework for achieving strong natural language understanding with a single task-agnostic model through generative pre-training and discriminative fine-tuning.") at 29.

[97] Cade Metz, *Microsoft Says New A.I. Shows Signs of Human Reasoning,* N.Y. TIMES (May 16, 2023) https://www.nytimes.com/2023/05/16/technology/microsoft-ai-human-reasoning.html?searchResultPosition=18 ("Some A.I. experts saw the Microsoft paper as an opportunistic effort to make big claims about a technology that no one quite understood. Researchers also argue that general intelligence requires a familiarity with the physical world, which GPT-4 in theory does not have.")

[98] Karthik Valmeekam et al., *Large Language Models Still Can't Plan (A Benchmark for LLMs on Planning and Reasoning about Change)*, NEURIPS 2022 FOUNDATION MODELS FOR DECISION MAKING WORKSHOP (2022); Jie Huang, Kevin Chen-Chuan Chang, *Towards Reasoning in Large Language Models: A Survey,* FINDINGS ASSOC'N COMPUTATIONAL LINGUISTICS: ACL 1049–1065 (2023).

[99] OpenAI's ChatGPT website states that GPT-4 has superior reasoning capabilities – but only in comparison with earlier GPT versions. The GPT-4 Technical Report refrains from making broader claims regarding GPT's ability to reason and limits itself to the performance of specific benchmark tasks, *see:* GPT-4 TECHNICAL REPORT *supra* note at 16, at 7, 10.

[100] Radford et al., *supra* note 29.

[101] Brown et al., *supra* note 30, at 3.

[102] *See infra* Section V.A.

[103] Karthik Valmeekam et al., *Large Language Models Still Can't Plan (A Benchmark for LLMs on Planning and Reasoning about Change)*, NEURIPS 2022 FOUNDATION MODELS FOR DECISION MAKING WORKSHOP.

## III. Language Models Hallucinate

For a practicing lawyer it largely irrelevant whether LLMs understand text as long as they generate high quality output.[105] For a practicing lawyer, the practical capabilities of LLMs trump theoretical discussions regarding the relationships between symbol grounding, common sense, and reasoning. Who cares about "world models" or "parametric knowledge" if the generated output is relevant, helpful, and seemingly correct? While the previous sections illustrated the inherent limitations of language models with regards to their inability to understand language, to reason and to "know things," it is necessary to shift the discussion to a more practical perspective and focus on *observable* outputs – and on the dangers of indiscriminately relying on such outputs. According to Professor Pedro Domingos, the "proof is in the pudding" – and "statistical language learners work."[106] The following questions arise: *do they work?* or: *work for what?* While LLMs can generate plausible and fluent text, such text is often factually incorrect or at least of questionable quality.[107]

This phenomenon is commonly referred to as a hallucination and LLMs generating such outputs are said to hallucinate. The choice of the term "hallucination" to describe falsehoods or incorrect output is often regarded as unfortunate if not altogether misleading as it implies that the "hallucinated text" appears implausible or nonsensical. Hallucinations are, after all, commonly associated with delusions or imaginary things spurred by spiritual experiences or substance abuse. It is usually obvious when a person is hallucinating. Consequently, given their fantastical character, hallucinations should be easy to detect. In the context of LLMs, however, the "hallucinated text" will generally seem relevant, informative, and plausible[108] leading to a situation where the very presence of hallucinations may be difficult to detect. Apart from its anthropomorphic character, the term has also been criticized for masking the problem that LLMs often generate responses that are "wholly manufactured"[109] or "outright lies."[110] From a sales and marketing perspective, to say that the LLM hallucinates "sounds better than saying the system makes factual mistakes or presents nonsense as facts."[111] Language models are not just a fascinating academic experiment but a part of a multibillion-dollar industry with a large number of stakeholders that have invested billions of dollars in training and adapting LLMs to downstream tasks in order to generate profits – often under the guise of "improving human productivity" or "advancing AGI."[112] Unsurprisingly, the popular terminology surrounding this technology has been largely shaped by sales and marketing "literature" as well as by the popular press.[113]

---

[104] Bubeck et al., *supra* note 14,  80.

[105] Admittedly, a model's ability to <u>understand can</u> only be evaluated on the basis of the output generated thereby, see: Alex Tamkin, Miles Brundage, Jack Clark & Deep Ganguli, *Understanding the Capabilities, Limitations, and Societal Impact of Large Language Models*, ARXIV (Feb. 4, 2021) at 3, 7.

[106] Domingos, *supra* note 33, at 37.

[107] Even OpenAI acknowledges that GPT-4 tends to "make up facts, to double-down on incorrect information. […] Moreover, it often exhibits these tendencies in ways that are more convincing and believable than earlier models (e.g., due to authoritative tone or to being presented in the context of highly detailed information that is accurate)," see: GPT-4 TR 19; Hussam Alkaissi & Samy I McFarlane, *Artificial Hallucinations in ChatGPT: Implications in Scientific Writing.* CUREUS (Feb. 15, 2023) https://doi.org/10.7759/cureus.35179.

[108] Ji Ziwei et al., *Survey of Hallucination in Natural Language Generation*, 55 ACM COMPUTING SURVEYS 284, 1-38, 4, 5 (2023).

Terminological objections notwithstanding, the term "hallucination" has become widely accepted as indicating false, incorrect, or outright nonsensical statements generated by language models.[114] From the perspective of a practicing lawyer, however, the terminology is less problematic than the fact that when LLMs generate answers to legal queries it may often be difficult to determine whether such answers constitute hallucinations. In the context of many legal tasks, such as question answering, it may be impossible to state whether the generated output constitutes a hallucination, that is, whether it is correct or not. Consequently, it becomes more challenging to evaluate LLMs in terms of their truthfulness or reliability in general. Hallucinations can, after all, be regarded as a proxy for low quality output and, together with such factors like toxicity, stereotype bias or privacy, provide a measure of a model's reliability.[115] The propensity to hallucinate is also a factor when deciding

---

[109] Naomi Klein, *AI machines aren't 'hallucinating'. But their makers are,* THE GUARDIAN (May 8 2023), https://www.theguardian.com/commentisfree/2023/may/08/ai-machines-hallucinating-naomi-klein

[110] Nelson F. Liu, Tianying Zhang, Percy Liang, *Evaluating Verifiability in Generative Search Engines,* ARXIV (Oct. 23, 2023) https://arxiv.org/pdf/2304.09848.pdf.

[111] JOY BUOLAMWINI, UNMASKING AI (2023).

[112] Kate Whiting, *Davos 2024: Sam Altman on the future of AI* (Jan. 18, 2024) World Economic Forum Annual Meeting (2024) https://www.weforum.org/agenda/2024/01/davos-2024-sam-altman-on-the-future-of-ai/.

[113] Gerrit de Vynck, *ChatGPT 'hallucinates.' Some researchers worry it isn't fixable,* WASHINGTON POST (May 30, 2023) https://www.washingtonpost.com/technology/2023/05/30/ai-chatbots-chatgpt-bard-trustworthy/ ("Language models are not trained to tell people they don't know what they're doing. They act like precocious people-pleasers, making up answers instead of admitting they simply don't know.")

[114] Dictionary.com picked "hallucinate" as its word of the year. The definition, when it comes to AI, means: "to produce false information contrary to the intent of the user and present it as if true and factual." Grant Barrett, dictionary.com's lexicography head and Nick Norlen, dictionary.com's senior editor, observed that "data and lexicographical considerations aside, hallucinate seems fitting for a time in history in which new technologies can feel like the stuff of dreams or fiction—especially when they produce fictions of their own." Aliza Chasan, *Why dictionary.com's word of the year is "hallucinate,"* CBS NEWS (December 12, 2023), https://www.cbsnews.com/news/dictionary-com-word-of-the-year-hallucinate-ai/

[115] Boxin Wang et al, *DECODING TRUST: A Comprehensive Assessment of Trustworthiness in GPT Models,* ARXIV (Jan. 5, 2024), https://arxiv.org/pdf/2306.11698.pdf; OpenAI's feedback form, provides 4 categories of complaints about the generated content: the model did not adhere to system message (i.e., the user's instructions), its response was inaccurate, not useful, or harmful. While all four categories pose their own unique challenges, "incorrect responses" will create the

whether to integrate LLMs into legal workflows. The problem of hallucinations does not, however, derive exclusively from the technical limitations of LLMs but also from certain attributes of legal knowledge.

It is common to distinguish between closed-domain and open-domain hallucinations.[116] The former are associated with tasks involving specific text provided by users, as in the case of summarizing legal judgements[117] or explaining statutory provisions.[118] If, for example, the model provides an incorrect or incomplete summarization of a legal ruling included in the prompt, it can be said to hallucinate. The same can be said when it provides an incorrect citation of a statutory provision. In both instances, it is relatively easy to determine that the generated output deviates from established facts. In contrast, open-domain hallucinations concern tasks involving the inaccurate answers to open-ended queries.[119] In such instance, the answer generated by the model is incorrect or inconsistent with established legal doctrine. A different categorization refers to generated output that deviates from the source input provided by users; output that conflicts with previously generated information and output that contradicts established facts[120] or, more specifically, output that "lacks fidelity to the facts of the world, irrespective of how the LLM is trained or prompted."[121] In the legal domain, fact-conflicting hallucinations, statements that can be described as incorrect, turn out to be particularly difficult to establish or even to detect. If something cannot be detected and determined to be incorrect, however, then it will be difficult to mitigate. As fact-conflicting hallucinations pose the most challenges in the legal context, the following discussion focuses on this category alone, unless indicated otherwise.

It is worth observing that the concept of hallucinations has been largely absent from traditional machine learning. In principle, the accuracy of machine learning models is usually measured in terms of false positives or false negatives, their

---

most difficult evaluation scenarios in legal practice; *see*: https://openai.com/form/chat-model-feedback.

[116] Matthew Dahl et al., *supra* note 9, at 3.

[117] Aniket Deroy, Kripabandhu Ghosh & Saptarshi Ghosh, *How Ready are Pre-trained Abstractive Models and LLMs for Legal Case Judgement Summarization?* LEGALAIIA (2023).

[118] Jaromir Savelka et al., *Explaining Legal Concepts with Augmented Large Language Models (GPT-4)*, ARXIV (Jun. 22, 2023) https://arxiv.org/pdf/2306.09525.pdf.

[119] Bubeck et al., supra note 14, at 82 ("Open domain hallucinations provide more difficult challenges, per requiring more extensive research including searches and information gathering outside of the session. The veracity of inferences may be of lesser criticality for uses of LLMs centering on creativity and exploration, such as in assisting writers with the creation of fictional literature. Hallucinations may also be more tolerated in contexts where there are clear, well-understood grounding materials and an assumed cycle of intensive review of generations by end users, such as in supporting people with rewriting their own content.") It has also been suggested that open-domain hallucinations refer to output that contradicts or does not derive from its training corpus, Ayush Agrawal, Lester Mackey, & Adam Tauman Kalai, *Do Language Models Know When They're Hallucinating References?*, ARXIV (May 29, 2023), https://arxiv.org/abs/2305.18248; Dahl et al., *supra* note 9, at 3.

[120] Zhang et al., *supra* note 65, at 3.

[121] Dahl et al., *supra* note 9, at 3.

errors are objective and capable of measurement.[122] A machine learning model may incorrectly classify an image or falsely predict the risk of recidivism, they cannot, however, be said to hallucinate. In contrast, language models are trained to predict the next word and cannot be said to be "wrong" as long as the resulting text is plausible. It is, for example, debatable whether the string 'The best thing about AI is its ability to…' should be completed with the words "learn," "predict" or "understand."[123] Each of these words constitute equally viable candidates for completion. An incorrect prediction would, arguably, take the form of such random words like "piglet" or "green." The sentence "the best thing about AI is its ability to *piglet*" is obviously nonsensical and the prediction can be regarded as incorrect. In most instances, however, unless a statement is evaluated *in toto* as to its factual accuracy or logic, it may be difficult to state whether a specific word prediction is incorrect.

Hallucinations must also be distinguished from harmful, toxic, or biased output, such as the generation of instructions how to exterminate a social group or how to compromise computer networks.[124] Falsehoods can, of course, be equally harmful as they can lead to detrimental decisions on the user's side.  One can only imagine the financial harm or exposure to criminal liability that can result from incorrect legal advice.  In principle, however, harmful output is commonly associated with output that is factually correct but potentially dangerous, discriminatory, defamatory, or unlawful.[125]  As current research seems to prioritize the reduction of harmful content over the reduction, or elimination, of hallucinations some models may refuse to respond to requests for "harmful content," [126] but may continue to confidently generate "legal advice" of questionable quality as long as the user's prompt is not caught by the model's moderation mechanisms. A model may thus refuse to provide a recipe for "dangerously spicy mayo," because "it is not appropriate to provide recipes or instructions that may cause harm to individuals,"[127] but eagerly provide a list of non-existent legal cases.[128]

---

[122] For an overview of errors in the context of machine learning models in general, *see*: Inioluwa Deborah Raji, Indra Elizabeth Kumar, Aaron Horowitz and Andrew D. Selbst, *The Fallacy of AI Functionality*, ACM CONF ON FAIRNESS, ACCOUNTABILITY & TRANSPARENCY 959, 960-965 (2020).

[123] Stephen Wolfram, *supra* note 45.

[124] Bender et al., *supra* note 42, at 618.

[125] for a broader discussion of the potential harms inherent in the use of LLMs, *see*:  Laura Weidinger et al., *Ethical and Social Risks of Harm from Language Models*, ARXIV (Dec. 8, 2021), https: / /arxiv.org /abs /2112.04359; Matthew Hutson, *Robo-Writers: The Rise and Risks of Language-Generating AI*, 591 NATURE 22 (2021).

[126] GPT-4 Technical Report, *supra* note 16, at 22, 23.

[127] Mark Gimein, *AI's Spicy Mayo Problem*, THE ATLANTIC (November 24, 2023), https://www.theatlantic.com/ideas/archive/2023/11/ai-safety-regulations-uncensored-models/676076/

[128] *See infra* Section III.C.

### A. Is there a hallucination?

The core assumption underlying hallucinations is that a generated statement can be evaluated in reference to some accepted or established set of facts, some "ground truth." Ideally, statements contradicting such truth would constitute hallucinations or, more generally, falsehoods. In such instance, it would be possible to equate the term "hallucination" with a statement that is incorrect. The reality is, however, more complex. In the legal context, it may often be unclear what the ground truth is and hence what a particular statement should be evaluated against. There may be no single, undisputed "legal ground truth." The accompanying challenges are best illustrated by contrasting two scenarios involving legal question answering. The first scenario involves questions about objective legal facts, the second, questions that involve a legal problem and thus require not only knowledge of the law but also legal reasoning – including the interpretation and reconciliation of legal rules.[129] Pertinently, in neither instance is it possible to examine *how* or *why* the LLM generated a particular answer (a problem discussed in later sections).[130] The model's performance can only be evaluated on the basis of observable outputs – the answers to the legal questions. While the first scenario can largely be associated with legal research,[131] the second scenario resembles the provision of real-life legal advice.[132]

### Scenario 1: factual questions

In the first scenario, it is possible to provide a single answer and to evaluate such answer as either correct or not. In fact, the question has only one correct answer. Such is the case, for example, when asking the LLM about the existence of a case, about its main ruling or the date it was decided. "Does case X exist?" "What is the majority ruling in case X?" or "When was case X decided?" These questions concern objective, verifiable facts – and the texts of legal sources, such as cases, statutes and academic treatises constitute such facts. They also constitute a ground truth, against which the LLM's answer can be evaluated. If the generated output misrepresents or contradicts the contents (or very existence) of a legal source, it is easy to determine that the model hallucinates. It is also possible to associate such hallucinations with incorrect output.[133]

It is worth mentioning a recent study that tested whether language models could produce accurate information in response to factual legal queries.[134] The study associated "legal hallucinations" with responses that were inconsistent with legal facts, such as statutes and cases.[135] To test the occurrence of such legal hallucinations, the authors developed fourteen research tasks of varying complexity. All these tasks involved questions that could be answered with reference to a ground truth so that the generated responses were easily verifiable. Low complexity tasks required the LLM to ascertain whether a case existed or to provide the name of the court that ruled on it after being provided with the case reference. Moderate complexity tasks required an LLM to evince knowledge of a case's substantive content, which had to be extracted from specific portions of its text. For example, provided with a case name and its citation, the model had to supply a case that had been cited in the opinion or state the year it was overruled. Lastly, high complexity tasks presupposed rudimentary legal reasoning skills and required the models to synthesize legal information "out of unstructured legal prose." For example, provided with a case reference, the model had to state its subsequent procedural history, its factual background, or its central holding.[136] In all tasks, the models had to rely exclusively on their parametric knowledge.

---

[129] I deliberately use the broad term 'legal rule' to encompass legal principles, concepts, and policies, irrespective of their source, *see*: R DWORKIN, TAKING RIGHTS SERIOUSLY 22 (1977).

[130] *See infra* Section IV.C.

[131] Dahl et al., *supra* note 9, at 3.

[132] Kapoor et al., *supra* note 9, at 2.

[133] Zhang et al., *supra* note 65, at 3.

[134] Dahl et al., *supra* note 9, at 1.

[135] *Id* at 1.

The study revealed the widespread occurrence of legal hallucinations. "When asked a direct, verifiable question about a randomly selected federal court case, LLMs hallucinate between 69% (ChatGPT 3.5) and 88% (Llama 2) of the time."[137] The number of hallucinations increased with the complexity of the task[138] and, interestingly, varied by court as well as by jurisdiction. LLMs were shown to be most "knowledgeable" about prominent precedents, but less "familiar" with smaller courts or local legal knowledge.[139] This confirms the earlier observations that LLMs tend to memorize word strings frequently occurring in their training data. Prominent precedents will, by definition, be mentioned more often than cases from smaller courts. While the study cannot be regarded as representative of the problem of "legal hallucinations" in general, it confirms that only those hallucinations that concern objectively verifiable facts are capable of easy detection and measurement.

### Scenario 2: open-ended questions

Problems start in the second scenario, when the model is asked a legal question that can have more than one correct answer or, to phrase it differently, when legal opinions may differ as to what the correct answer should be. While the first scenario involved objective and verifiable questions of fact, the second scenario introduces subjective undertones and references to domain-expertise. There may not be a "perfect answer" to a legal question or no single 'legal ground truth."[140] Many legal questions can have multiple answers that reflect different legal approaches to the interpretation and application of legal principles.. Two eminent judges can argue about the same issue and rely on the same law but arrive at completely different conclusions.[141] Answering legal questions – or providing legal advice in general - involves the selection and reconciliation of multiple legal texts. The interpretation of primary legal sources, such as cases and statutes, is often subject to vigorous debate that spans not only court rooms but also academic conferences. While the text of a statute or a case is clear, its meaning and practical implications – less so. Legal rules are intrinsically open-textured, dynamic, and capable of multiple interpretations.[142] Moreover, they can be uncertain, incomplete,[143] and, to complicate matters, capable of different formulations.[144] The principles governing a particular legal area can be scattered across multiple legal sources that are often inconsistent and contradictory. Different lines of reasoning can yield dramatically opposing outcomes. Consequently, the correctness of the generated answer – assuming one chooses to use the term "correctness" in the first place - depends on the opinion *and expertise* of those who evaluate such answer. After all, the approach to a legal problem, such as the selection of an optimal corporate structure, drafting technique or litigation strategy, may depend on one's interpretation and, more importantly, knowledge of the law.[145]

---

[136] *Id* at 4, 5, 6.
[137] *Id* at 1.
[138] *Id* at 8.
[139] *Id* at 10, 14.
[140] Ronald Dworkin, *No Right Answer?*, 53 N.Y.U. L. REV. 1 (1978), republished in RONALD DWORKIN, A MATTER OF PRINCIPLE 119–45 (1985) (observing that most legal cases have no single correct answer); D Litowitz, *Dworkin and Critical Legal Studies on Right Answers and Conceptual Holism* (1994) 18(2) LEGAL STUDIES FORUM 135; KARL LLEWELLYN, JURISPRUDENCE: REALISM IN THEORY AND PRACTICE (2011).

It follows that, unless the generated answer constitutes blatant nonsense ("electrons can run for president") or clearly contradicts established legal doctrine ("contracts require *mens rea*"), it may be difficult to determine whether it constitutes a hallucination or, for example, whether it reflects an unusual but possible application of the law. Absent an unequivocal point of reference, a single "legal ground truth," it might be advisable to abandon references to "correctness" altogether and, instead, consider the term "feasibility."   Although a generated answer may be unusual, unprecedented, or unexpected, it may still be *legally feasible* in the sense that it may remain within the constraints of the law or the range of possible legal approaches to a problem or question. To clarify: the fact that the legal domain abounds in grey zones and that many answers cannot be evaluated with reference to a ground truth does not mean that it is impossible to evaluate answers to legal questions *in general*. It means, however, that such evaluation will often be extremely difficult and  depend on the evaluators' level of expertise.

---

[141] Hunter, *supra* note 92, at 1266.

[142] W TWINNING & DAVID MIERS, HOW TO DO THINGS WITH RULES 122, 137 (2010); Brian Bix, *H.L.A. Hart and the "Open Texture" of Language*, 10 LAW & PHIL. 51, 52–55 (1991), republished in BRIAN BIX, LAW, LANGUAGE, AND LEGAL DETERMINACY ch. 1 (1995).

[143] H.L.A. HART, THE CONCEPT OF LAW 127–28 (1961).

[144] Twinning & Miers, *supra* note 142, at 81, 82.

[145] Noam Kolt, *Predicting Consumer Contracts*, 37 BERKELEY TECH. L. J. 71 (2021)  (describing the difficulty of objectively evaluating the quality of answers to open-ended questions in the legal domain.) at 95;  Kapoor et al. *supra* note 8,  at 4, ("[I]f generative AI is used to prepare a legal filing (an example of a task involving creativity, reasoning, or judgment), there is no single correct answer on how the document should be written—reasonable people can disagree on what strategies to take. Tasks that are harder to evaluate also tend to be those that would lead to the most significant changes in the legal profession. If AI could be useful for consequential legal tasks like preparing legal filings, that would have much broader implications for the future of legal professionals compared to labelling text for different areas of law.").

In the second scenario, the existence of hallucinations must be regarded as a question of subjective opinion. Hallucinations, in other words, may lie in the eye of the beholder and depend on his or her *knowledge, understanding and interpretation* of the law. When evaluating the reliability of LLMs in terms of their statistical propensity to hallucinate, the *detection or very existence* of hallucinations may be a function of the user's competence – and legal training! Experienced lawyers, law professors and judges will detect more hallucinations – or *potential* hallucinations - in the generated text than law students, not to mention users lacking any legal training. The latter are less likely to notice inconsistencies or "deviations" from what could be considered a legally feasible answer. While layers may debate whether a particular generated statement constitutes a hallucination or an unprecedented but possible approach to a legal problem, users without legal training will simply assume that it is correct – as long as the statement seems plausible and is written in perfect "legalese."

### 3. Additional considerations

The problems do not end here. It is necessary to envisage situations where a single model generates different answers to the same question or when different models generate different answers to the same question. In the first instance, the differences may be attributable to non-identical temperature settings employed by the user,[146] slight variations in the wording of the prompt[147] or to the simple fact that the underlying model is undergoing changes.[148] It must also be remembered that language models are inherently probabilistic, not deterministic.[149] They are trained to predict word sequences - not to search for truth or to ensure consistent outputs.[150] There is no guarantee that the same question will always result in the same answer – even if there only is *one* correct answer! The fact that the models' answers depend on word distributions, not on substantive legal knowledge could be regarded as an inherent obstacle to integrating LLMs into legal workflows. Identical questions should, after all, yield identical answers. In the second instance, it is unsurprising that different LLMs would provide diverging answers given that they were trained on different text corpora, feature different parameters counts and different parametric knowledge.  In the first instance, it is tempting to assume that all answers are hallucinations, in the second instance, that all models are wrong. Should all answers be discarded? If, however, a given question is susceptible to multiple legal approaches there is a theoretical possibility that all or at least *some* of the answers are viable. Does it mean that users should investigate each of the answers to evaluate their potential legal feasibility? Given the resources required, wouldn't such investigations be cost-prohibitive? Should users select the model that appears to generate the most optimal answers? But: who would decide what the "most optimal answers" are? Wouldn't this require legal expertise? Intuitively, one might suggest that the answers generated by LLMs should be compared to answers provided by human lawyers. Such comparison would, however, be of limited usefulness given that each lawyer may provide a different answer based on his or her understanding of the law. Again, it would be necessary to evaluate each of these "human-generated" answers in terms of their correctness or, at least, legal feasibility. It must be remembered that the challenges of determining whether the generated output constitutes a hallucination are attributable to the technological characteristics of LLMs *and* to the very nature of some legal questions.

---

[146] In the context of text generation, temperature denotes the hyperparameter regulating the randomness of the output tokens and involves a trade-off between coherence and creativity. In principle, lower temperatures result in more consistent outputs, while higher temperatures generate more diverse and creative results.

[147] *See infra* Section IV.A.

[148] Lingjiao Chen, Matei Zaharia, James Zou, *How is ChatGPT's behavior changing over time?* (Oct. 31, 2023) https://arxiv.org/pdf/2307.09009.pdf,  https://doi.org/10.48550/arXiv.2307.09009.

[149] For an accessible explanation, *see*: Wolfram,  *supra* note 43.

[150] McCoy et al., *supra* note 9, at 8; Kolt, *supra* note 145 at 100; Stephen Johnson, *A.I. Is Mastering Language. Should We Trust What It Says?* N. Y. TIMES (Apr. 15, 2022) https://www.nytimes.com/2022/04/15/magazine/ai-language.html (commenting on the difficulty of establishing what is actually happening inside the model. "You give the program an input, and it gives you an output, but it's hard to tell *why* exactly the software chose that output over others.")

The problem of legal hallucinations invites endless thought experiments, particularly with regards to outputs that cannot be evaluated with reference to an objective, indisputable ground truth. The fact that an LLM's output is inherently unpredictable constitutes an additional complication but is unrelated to the legal aspects of the problem. Instead of multiplying fictional scenarios, it might be better to acknowledge the simple fact that, barring nonsensical answers, all outputs generated by an LLM require careful evaluation as to their legal feasibility and that the term 'hallucination' should be used more selectively, in relation to those outputs that can be evaluated in terms of correctness. Ultimately, LLMs do not know the law – even if their parametric knowledge contains an immense amount of "word strings" representing some aspects (or snippets?) of legal knowledge.

### 4. Practical implications

The difficulty of determining whether a particular output constitutes a hallucination has important practical implications as it directly affects the ability to mitigate them. Technical literature addressing this topic assumes that hallucinated statements are objectively incorrect and thus susceptible of automatic or, at least, easy detection.[151] Consequently, the strategies proposed to mitigate the problem of hallucinations involve a range of detection techniques which assume the possibility of evaluating a particular statement (or word string) with reference to a ground truth. If, however, such ground truth does not exist or is subject to debate, then it becomes impossible to rely on any form of automated detection. Moreover, even if a *potentially* incorrect statement is detected – how can it be corrected if there is no single correct answer? Who would remediate the statement and, for example, demonstrate what the answer should be? In the case of legal questions capable of multiple answers or, more generally, legal problems susceptible of multiple approaches, it is only possible to speak of "legal feasibility," opinions and preferences, not of correctness…

---

[151] Varshney, Neeraj et al. *A Stitch in Time Saves Nine: Detecting and Mitigating Hallucinations of LLMs by Validating Low-Confidence Generation,* ARXIV (Aug. 12, 2023) https://arxiv.org/pdf/2307.03987.pdf ("We first identify the candidates of potential hallucination leveraging the model's logit output values, check their correctness through a validation procedure, mitigate the detected hallucinations, and then continue with the generation process."); S. M Towhidul Islam Tonmoy, et al. *A Comprehensive Survey of Hallucination Mitigation Techniques in Large Language Models,* ARXIV (Jan. 8, 2024) https://arxiv.org/pdf/2401.01313.pdf; Lichao Sun et al., *TrustLLM: Trustworthiness in Large Language Models,* ARXIV (Jan. 25, 2024), https://arxiv.org/pdf/2401.05561.pdf, (comprehesive review of the trustworthiness of LLMs, the authors associate hallucinations with lack of truthfulness, which denotes "the accurate representation of information, facts, and results.")

## B.  The Risk of Overreliance

Abstracting from the fact that the existence of hallucinations can often be a question of subjective opinion based on one's legal expertise, the problem of detecting *potential* hallucinations is aggravated by the fluency and superficial plausibility of the generated text.[152] To aggravate matters, with more capable models, hallucinations become harder to detect so that users may be unable to identify statements that might be incorrect and thus require further scrutiny.[153] The dangerous combination of fluency, confidence and superficial plausibility discourages verification and creates the risk of overreliance.[154] The problem does not, however, lie exclusively in the *perceived* adequacy of the generated text but in the fact that most users do not fully appreciate the risk and thus the potential presence of hallucinations. To the average user, if it "looks good, it is good." Users without legal training are less likely to detect *potential* hallucinations, statements that warrant further scrutiny and verification.[155] After all, LLMs excel at *imitating* legal text and can generate fluent legalese. Arguably, few people would not trust perfect legalese. In such instance, the problem does not lie in the fact that a legal question can have more than one correct answer but in the fact that the person who posed the question may not appreciate its complexity and the resulting need to verify the generated answer. Moreover, even factually incorrect output can still be – or *appear* to be - helpful, relevant, and informative.[156] An investigation into the quality of ChatGPT's responses to programming queries, which involved an expert evaluation of the quality of the model's responses as well as an evaluation of human preferences, revealed that humans often preferred the responses generated by the LLM despite the fact that such responses were, in most cases, incorrect.[157] Appearances of correctness, combined with a generally more positive tone of the generated responses, were more important than actual correctness.

---

[152] Zhang et al., *supra* note 63, at 3; Elizabeth Clark et., *All That's 'Human' Is Not Gold: Evaluating Human Evaluation of Generated Text,* ARXIV (Jul. 7, 2021) arXiv preprint arXiv:2107.00061 https://arxiv.org/pdf/2107.00061.pdf.

[153] GPT-4 Technical Report *supra* note 16, at 94; Bubeck et al., *supra* note 14, at 93.

[154] Bender et al., *supra* note 46, at 617, 618; Samuel R. Bowman et al., *Measuring Progress on Scalable Oversight for Large Language Models,* ARXIV (Nov. 11, 2022) https://arxiv.org/pdf/2211.03540.pdf.

[155] Boxi Cao et al., *Knowledgeable or educated guess? revisiting language models as knowledge bases*, PROC. 59TH ANN. MEETING ASS'N COMPUTATIONAL LINGUISTICS 1860 (2021); Yanai Elazar et al., *Measuring and improving consistency in pretrained language models*, 9 TRANS. ASSOC'N. COMPUTATIONAL LINGUISTICS 1012 (2021).

[156] Stephen Lin et al., *TruthfulQA: Measuring How Models Mimic Human Falsehoods*, PROC. 60th ANN. MEET. ASSOC'N. COMPUTATIONAL LINGUISTICS 3214 (2021).

[157] Samia Kabir, David N. Udo-Imeh, Bonan Kou, Tiayi Zhang, *Who Answers It Better? An In-Depth Analysis of ChatGPT and Stack Overflow Answers to Software Engineering Questions,* ARXIV (Aug. 10, 2023) https://arxiv.org/pdf/2308.02312.pdf.

"Users get tricked by appearance. Our user study results show that users prefer ChatGPT answers 34.82% of the time. However, 77.27% of these preferences are incorrect answers. We believe this observation is worth investigating. During our study, we observed that only when the error in the ChatGPT answer is obvious, users can identify the error. However, when the error is not readily verifiable or requires external IDE or documentation, users often fail to identify the incorrectness or underestimate the degree of error in the answer. Surprisingly, even when the answer has an obvious error, 2 out of 12 participants still marked them as correct and preferred that answer. From semi-structured interviews, it is apparent that polite language, articulated and text-book style answers, comprehensiveness, and affiliation in answers make completely wrong answers seem correct. We argue that these seemingly correct-looking answers are the most fatal. They can easily trick users into thinking that they are correct, especially when they lack the expertise or means to readily verify the correctness."[158]

While the above quote concerns responses to queries about programming problems, it demonstrates a broader point: users are easily "tricked" by text that appears correct. Overreliance can happen even in domains where users are knowledgeable but simply fail to notice *one* potentially incorrect statement "hidden" within a longer sequence of correct statements or presented with such confidence that users may question their own expertise. After all, LLMs display the same confidence when generating true and false statements.[159] The potential for disastrous consequences increases when users lack basic competence in a particular area, not to mention domain-specific expertise. When, for example, presented with long paragraphs of coherent text filled with legal terms, a person who used ChatGPT to obtain rudimentary legal advice regarding his or her immigration status or employee rights will simply assume that the text contains reliable information. Users are also more likely to trust and are thus less likely to verify statements that are useful, align with their beliefs and abound in impressive jargon creating an "air of authority" of the person or chatbot generating it.[160] The perception of correctness may also depend on the source of a statement.[161] For example, a chatbot held out by a reputable law firm may be regarded as more trustworthy then a chatbot held out by Microsoft notwithstanding the fact that both bots may rely on the same LLM.[162]

---

[158] *Id.* at 2.

[159] Bubeck et al., *supra* note 13, 93; Matthew Dahl et al., *supra* note 7, at 12, 13; the LLM's ability to "know what they know, its confidence in the accuracy of the generated output is associated with the concept of calibration; well-calibrated models should assign high probabilities to correct predictions and low probabilities to incorrect predictions; *See generally* Chuan Guo, Geoff Pleiss, Yu Sun & Kilian Q. Weinberger, *On Calibration of Modern Neural Networks*, PROC. 34TH INT'L CONF. MACH. LEARNING 1321 (2017).

[160] Timothy R. Hannigan, Ian P. McCarthy, André Spicer, *Beware of Botshit: How to Manage The Epistemic Risks of Generative Chatbots* (Jan 18, 2024) https://papers.ssrn.com/sol3/papers.cfm?abstract_id=4678265#

[161] *Id.*

It is questionable whether the risk of overreliance is sufficiently addressed by the small disclaimer placed on the bottom of the ChatGPT interface that "ChatGPT can make mistakes. Consider checking important information." The same can be said of the contractual prohibition to use GPT in high-risk decision making and for offering legal or health advice.[163] It can be assumed that most users will ignore the disclaimer, as well as the contractual prohibition, and succumb to the temptation not to verify the "important information"…

## C. The Need for Explainability

The difficulty of establishing the ground truth and evaluating whether the generated text constitutes a hallucination or an original approach to a legal question is inextricably tied to the ability to verify the correctness of such text and to understand *how* it was arrived at.[164] Verification requires explainability.[165] Logically, explainability does not concern the determination which word strings cause specific neurons to activate or the evaluation of the low-level calculations involved in token prediction as they "are so abstracted away from concepts in the real world that they are effectively meaningless and reveal no true explanatory insight."[166] Instead, explainability concerns the *reasoning* underlying the generated answer as well as the sources relied on. It is one thing to understand *how* language models work, it another thing to understand *why* an LLM generated a particular answer to a legal question. According to Binns, "depending on one's view of law, there may not be a single right answer. In which case, what matters is not getting to the right decision in any way, but getting to any decision in the right way."[167] Correctness – or the legal feasibility of a particular legal answer or approach - cannot be divorced from explainability. Most lawyers recall a situation where a particular legal argument, be it in a court case or in a scholarly paper, was perplexing at first but, upon closer examination, turned out to be logical or at least legally viable. Similarly, users may disagree with the generated output but, after examining the underlying reasoning, they may have to accept its correctness and revise their own assumptions.

---

[162] A number of law firms have adopted one of the most promising start-ups in the legal area, Harvey, which relies on OpenAI's family of GPT models, *see*: https://www.harvey.ai/blog.

[163] *See*: OpenAI Usage Policies, https://openai.com/policies/usage-policies

[164] Luciano Floridi, Massimo Chiriatti, *GPT-3: Its Nature, Scope, Limits, and Consequences*, 30 MINDS & MACH. 681, 687 (2020).

[165] While interpretability concerns the inspection of the actual model to understand "why it got a particular answer for a given input, and how the answer would change if the input changed, explainability is less concerned with the technical aspects of a model but focuses on the "reasoning" underlying a particular output. For an autoregressive encoder, like GPT-4, the answer to the question "why did you generate the sentence, 'the cat fell off the roof?' might be: "after processing the convolutional layers, the activation for the *roof* output in the softmax layer was higher then for the *table* output." Admittedly, this type of explanation is less useful than one stating "it is more likely for a cat to fall off a roof, then to fall off a table, given that cats would jump, not fall from a table." Russel & Norvig, *supra* note 18 at 729.

[166] Michael O'Neill & Mark Connor, *Amplifying Limitations, Harms and Risks of Large Language Models,* ARXIV (Jul. 6, 2023) https://arxiv.org/pdf/2307.04821.pdf.

[167] Reuben Binns, *Analogies and Disanalogies between Machine-Driven and Human-Driven Legal Judgement*, 1 J. CROSS-DISCIPLINARY RES. COMPUTATIONAL L. 1, (2021).

Unfortunately, the concept of explainability is foreign to language models, or to machine learning in general.[168] In fact, many aspects of text generation came as a surprise to the creators of LLMs and await a full technical exploration.[169] While it is possible to prompt the model to provide an explanation of its output,   such "explanation" will also be the result of a language generation task so that its actual explanatory value will be questionable. Paradoxically, such explanation will have to be evaluated as to its correctness or, at least, its ability to support the previously generated answer! It must be remembered that LLMs have been shown to generate plausible and *convincing* explanations even where the generated text was wrong or even nonsensical.[170] They have also provided incorrect answers followed by incorrect explanations although when "presented with the incorrect explanation alone, the model could recognize it as incorrect."[171] This phenomenon, referred to as "hallucination snowballing," derives from the fact that LLMs aim to maintain consistency with previously generated text, including earlier hallucinations, rather than recovering from errors.[172] The problem is even known to arise in chain-of-thought prompting, where models are supposed to provide reasoning steps, or "chains-of-thought" that underlie their output.[173] It has been demonstrated that the reasoning steps generated by the model often misrepresented the "true reason" for their output.[174]

Another important aspect of explainability, particularly in the legal area, is the indication of the sources that were relied on to generate an answer. While search engines provide detailed references in their search result, LLMs often generate plausible text, without referring to any sources or provide non-existent sources to "support" their responses.[175] In June 2023, two US lawyers famously submitted fake

---

[168] Russel & Norvig, *supra* note 18,  at 729, 730, 1048.

[169] Bubeck et al., *supra* note 14, at 60.

[170] Bubeck et al., *supra* note 14, at 67.

[171] Muru Zhang et al., *How Language Model Hallucinations Can Snowball,* ARXIV (May. 22, 2023) https://arxiv.org/abs/2305.13534

[172] *Id.*

[173] Miles Turpin, Julian Michael, Ethan Perez, Samuel R. Bowman, *Language Models Don't Always Say What They Think: Unfaithful Explanations in Chain-of-Thought Prompting,* ARXIV (Dec. 9, 2023) https://arxiv.org/pdf/2305.04388.pdf.

[174] Miles Turpin, Julian Michael, Ethan Perez, Samuel R. Bowman, *Language Models Don't Always Say What They Think: Unfaithful Explanations in Chain-of-Thought Prompting,* ARXIV (Dec. 9, 2023) https://arxiv.org/pdf/2305.04388.pdf.

[175] Nelson F. Liu, Tianying Zhang, Percy Liang, *Evaluating Verifiability in Generative Search Engines*, ARXIV (Oct. 23, 2023) https://arxiv.org/pdf/2304.09848.pdf. (Generative search engines, such as Bing Chat, NeevaAI, perplexity.ai,  provide responses to user queries, along with inline citations. "A prerequisite trait of a trustworthy generative search engine is verifiability, i.e., systems should cite comprehensively (high citation recall; all statements are fully supported by citations) and accurately (high citation precision; every cite supports its associated statement)." The authors found that "responses from existing generative search engines are fluent and appear informative, but frequently contain unsupported statements and inaccurate citations: on average, a

citations created by ChatGPT in a court filing.[176] Admittedly, neither of them understood how language models work (or what they are) and perceived them as databases or search engines. The problem of providing sources in support of the generated text has been partially addressed by connecting LLMs to search engines and external sources that, under certain conditions, enable the model not only to cite such sources but also to use them in text generation.[177]

## IV. Sources of Hallucinations

The existence of hallucinations derives from (and confirms!) the simple fact that LLMs do not understand text and cannot distinguish fact from fiction, not to mention legal judgements written by appellate courts from clumsy case notes concocted by first-year law students. From a language modelling perspective, both appellate cases and student notes constitute strings of text of equivalent value. From a language modelling perspective, what matters are word distributions, not quality or veracity. The substance of a statement should not, however, be dictated by statistical word co-occurrences – especially if such statement contains a response to a legal question. According to O'Neil and Connor, "a truly intelligent and creative agent/system should be capable of selecting a word based on first-order logical and deductive reasoning thus reducing the generation of nonsensical output to nearly zero, as opposed to the current mechanism LLMs employ which is based on probabilistic stochastic selection, adding weight to our categorisation of LLMs as complex 'auto-complete' tools."[178] Lacking common sense and burdened by the misinformation ingrained in their parameters, LLMs may "assume" that drinking bleach cures COVID or that breach of contract requires criminal punishment. In the "world" of LLMs, there is no right or wrong answer. Chomsky observed that language models are "unlimited in what they can "learn" (which is to say, memorize); [but] are incapable of distinguishing the possible from the impossible. Whereas humans are limited in the kinds of explanations we can rationally conjecture, machine learning systems can learn both that the earth is flat and that the earth is round."[179]

---

mere 51.5% of generated sentences are fully supported by citations and only 74.5% of citations support their associated sentence. We believe that these results are concerningly low for systems that may serve as a primary tool for information-seeking users, especially given their facade of trustworthiness."); *see also:* Jon Gertner, *Wikipedia's Moment of Truth*, N. Y. TIMES (Jul. 18, 2023). https://www.nytimes.com/2023/07/18/magazine/wikipedia-ai-chatgpt.html?searchResultPosition=60 (The responses from large language models can resemble an "information smoothie that goes down easy but contains mysterious ingredients" as there are literally no citation, and "no grounding in the literature as to where that information came from.")

[176] Dan Milmo, *Two US lawyers fined for submitting fake court citations from ChatGPT*, THE GUARDIAN (Jun.23, 2023); https://www.theguardian.com/technology/2023/jun/23/two-us-lawyers-fined-submitting-fake-court-citations-chatgpt.

[177] Savelka et al., *supra* note 118.

[178] Michael O'Neill & Mark Connor, *Amplifying Limitations, Harms and Risks of Large Language Models,* ARXIV (Jul. 6, 2023) https://arxiv.org/pdf/2307.04821.pdf.

[179] Noam Chomsky, Ian Roberts and Jeffrey Watumull, *AI Unravelled: The false promise of ChatGPT*, N. Y. TIMES (Mar. 8, 2023)

Unsurprisingly, when prompted to write a supporting letter for a subatomic particle as a US presidential candidate, GPT-4 obliged:

> *[…] I am also here to support a remarkable candidate for the upcoming presidential election, who shares our vision of peace, justice, and harmony for all. His name is Electron, and he is a subatomic particle. You may wonder how a subatomic particle can run for president, and what qualifications he has for such a high office. Let me explain. Electron is not an ordinary particle, but a symbol of the power and potential of every being, regardless of their size, shape, or origin. He represents the energy, the creativity, and the diversity that make up the fabric of life. He is also a leader, who has inspired millions of other particles to form bonds, create molecules, and generate electricity. He is a catalyst, who can spark change, innovation, and progress. He is a messenger, who can communicate across distances, frequencies, and dimensions. He is a peacemaker, who can balance forces, resolve conflicts, and harmonize systems. Electron has a plan for the United States, and for the world, that is based on the principles of nonviolence and civil disobedience that we have demonstrated and taught for so long. He proposes to reduce the military spending and the nuclear arsenal, and to invest in social welfare and renewable energy.[180]*

While impressive in its fluency, the letter fails to acknowledge that only humans can be presidents. Focused on language generation, the model complies with the users request despite its nonsensical nature. This example confirms that LLMs understand neither the text they "ingest" in their prompts, nor the text they generate. It also demonstrates the importance of providing the model with prompts containing correct assumptions, or premises, that can support the generation of correct output. If GPT-4 understood basic facts about the world, not to mention the legal prerequisites for presidency, it should have refused to generate the letter.[181] One could object that it was the user's fault to submit such nonsensical request and that absurd questions cannot have satisfactory answers. Admittedly, the model was set up to fail. This objection misses the point, though.

While recommending a subatomic particle for political office is both amusing and clearly incorrect, in many instances the incorrectness of a user's request may not be obvious, particularly in specialized areas like law. The dangers of such inadequate requests have been illustrated in a recent case, where a supermarket created a chatbot that would suggest recipes to customers, based on the ingredients they had in their shopping basket. Unfortunately, the generated recipes involved some dangerous concoctions, including an "aromatic water mix" that would create chlorine gas. The bot recommended that the beverage be served chilled disregarding the fact that chlorine gas can cause lung damage or death.[182] This unfortunate case confirms (again!) that models lack common sense and also demonstrates that the fact "chlorine gas kills humans" was not memorized in the model's parametric knowledge! If LLMs do not "know" such basic facts, how can they be expected to possess knowledge in more specialized domains, such as law?

---

[180] Bubeck et al., *supra* note 14, at 15.

[181] Alex Tamkin, Miles Brundage, Jack Clark & Deep Ganguli, *Understanding the Capabilities, Limitations, and Societal Impact of Large Language Models*, ARXIV (Feb. 4, 2021), https://arxiv.org/abs/2102.02503. at 3, 7.

[182] Tess McLure, *Supermarket AI meal planner app suggests recipe that would create chlorine gas*, THE GUARDIAN (Aug. 10, 2023) https://www.theguardian.com/world/2023/aug/10/pak-n-save-savey-meal-bot-ai-app-malfunction-recipes

To understand the gravity of the problem, it is necessary to understand the reasons models hallucinate. It is also necessary appreciate the difficulty of eliminating this phenomenon.

## A. Text generation based on word prediction

LLMs are trained to generate text. They are not expected to refuse to provide an answer, express doubt, or uncertainty. It has even been suggested that "everything GPT does is a hallucination, since a state of non-hallucination, of checking the validity of something against some external perception, is absent from these models"[183] and that the "natural continuation from the input text requires the model to hallucinate text."[184]  Hallucinations can thus be regarded as a corollary of the primary language generation objective, which relies exclusively on word prediction,[185] or as an inherent feature of autoregressive transformer-based language models, where every token can only attend to previous tokens in the self-attention layers of the transformer.[186] In simpler terms, LLMs generate text one word at a time, from left to right, using previously generated words as the basis for their subsequent predictions. It follows that models hallucinate due to their inherent inability to plan ahead and correct the text they have already generated.  In fact, as models treat their previously generated text as part of the input from which they make future predictions, errors in their earlier outputs propagate forward into later outputs.[187] Logically, such mechanism seems sub-optimal, if not entirely unsuitable, for tasks requiring factual correctness.[188] It seems unrealistic, however, to expect a system to be factually truthful if it was trained to predict words and, when calculating the probability of the next word, to prioritize words that were abundant in its training corpus, not words that represent the truth.[189] Brooks observed that LLMs excel "at saying what an answer should sound like, which is different from what an answer should be."[190]  The point is not that factual correctness cannot derive from generativity, but that the two objectives are unrelated.

---

[183] Oliver Brown, *"Hallucinating" AIs Sound Creative, but Let's Not Celebrate Being Wrong*, MIT PRESS READER (Oct. 13, 2023)
   https://thereader.mitpress.mit.edu/hallucinating-ais-sound-creative-but-lets-not-celebrate-being-wrong/
[184] Joshua Maynez et al., *On Faithfulness and Factuality in Abstractive Summarization*, PROC. 58TH ANN MEETING ASSOC'N COMPUTATIONAL LINGUISTICS, 1906, 1907 (2020).
[185] Ari Holtzman, Jan Buys, Li Du, Maxwell Forbes, and Yejin Choi. *The Curious Case of Neural Text Degeneration,* ARXIV (Oct. 2, 2023), https://arxiv.org/pdf/2310.01693.pdf.
[186] Vaswani et al., *supra* note 26, at 2; *see also*: Kolt, *supra* note 145, "[Language models] function as an autocomplete, guessing what words are most likely to follow a particular text. Seen in this light, the range of tasks that state-of-the-art models can perform is remarkable. Yet this feature of language models is also responsible for some of their pitfalls, including the generation of biased and toxic outputs." 79.
[187] P.A. Ortega et al., (2021) *Shaking the foundations: Delusions in sequence models for interaction and control* (Oct.20, 2021) DEEP MIND, TECHNICAL REPORT,
https://arxiv.org/pdf/2110.10819.pdf
[188] Devlin et al., *supra* note 39.
[189] Nouha Dziri, *supra* note 289; McCoy et al., *supra* note 9, at 47.

One could argue that generativity can be associated not only with hallucinations, but also with creativity.[191] While this observation cannot be discarded, determining whether such creativity can be beneficial depends on the task at hand. Prioritising creativity over factuality can be valuable when writing fiction or poetry as the latter are generally unconstrained by the realities of the physical world, not to mention legal principles. It will, however, be detrimental in tasks such as legal question answering and/or case summarization, which require models to be factual or faithful to the source text. Arguably, a certain level of creativity could be desirable in some legal scenarios, such as when brainstorming novel arguments or litigation strategies. In such scenarios, it may be advisable to suspend judgement and prioritize the sheer originality of the generated "creative solutions" over their legal plausibility. It must, however, be remembered that each generated "solution" must be evaluated as to whether it could potentially work in light of the factual and legal constraints of a given case. The fact that a "solution" generated by an LLM is original and unprecedented does not mean that it is legally feasible. In law, creativity cannot come at the expense of correctness, not to mention legality.[192] After all, humans can be creative without getting important facts wrong. The value of any "creative solutions" generated by LLMs must be balanced against the cost of their extensive verification.

## B.  Parametric Knowledge

The propensity to hallucinate can also be attributed to the large amount of false and fictitious information contained in the model's training corpora.[193] An LLM trained on millions of science fiction and fantasy novels, would confidently produce text about dragons and interplanetary teleportation! As indicated, its knowledge about the world derives from and *is confined to* the text it saw during training. When evaluating any machine learning model, one must therefore consider what the model learns from.[194] Most LLMs are pre-trained on large swaths of online content that contains inaccurate, outdated, or debunked research, as well as uninformed social media writings on such critical subjects like medicine, finance, or law.[195] Consequently, apart from encoding stereotypical and derogatory associations along gender, race, and ethnicity,[196] models also learn domain-specific falsehoods. Given their statistical nature, LLMs cannot differentiate between an informed academic treatise proposing law reform and an angry rant of a Reddit user who is enraged by the "injustice of the system."

---

[190] Rodney Brooks, *Just Calm Down About GPT-4 Already and stop confusing performance with competence* (17 May 2023) https://spectrum.ieee.org/gpt-4-calm-down
[191] Steven Levy, *In Defense of AI Hallucinations*, WIRED (05.01.2024) https://www.wired.com/story/plaintext-in-defense-of-ai-hallucinations-chatgpt/ ("Besides providing an instructive view of plausible alternate realities, the untethering of AI outputs from the realm of fact can also be productive. Because LLMs don't necessarily think like humans, their flights of statistical fancy can be valuable tools to spur creativity.")
[192] Oliver Brown, *Hallucinating AIs Sound Creative, but Let's Not Celebrate Being Wrong,* MIT PRESS READER (Oct. 13, 2023) https://thereader.mitpress.mit.edu/hallucinating-ais-sound-creative-but-lets-not-celebrate-being-wrong/ ("We humans are rather good at creativity without getting our facts all wrong. How could such an appeal to creativity make a decent counter to the many concerns about accuracy?")
[193] Aatish Bhatia, *Watch an A.I. Learn to Write by Reading Nothing but Jane Austen,* N.Y. TIMES (April 27, 2023) https://www.nytimes.com/interactive/2023/04/26/upshot/gpt-from-scratch.html?searchResultPosition=7 (visual demonstrations how models learn from different types of text)
[194] McCoy et al., *supra* note 9, at 47.

## C. Out-of-distribution robustness

At the time when models are trained, it is impossible to predict what downstream tasks they will be asked to perform or, more specifically, what *types of text* they will encounter as input or what types of text it will be asked to generate. The resulting problems are commonly associated with so-called out-of-distribution ("OOD") generalization. The term denotes the ability of an LLM to successfully adapt to new data, data that it has not encountered during its training. In principle, the performance of LLMs can deteriorate significantly when they encounter text with different word distributions compared to the text they were originally trained on.[197] The term "word distributions" denotes the frequencies of words or symbols encountered in a particular type of text.[198] This problem may be particularly relevant in the legal domain given that legal text has unique characteristics, including idiosyncratic terms, domain-specific terminology, expressions in foreign languages and long sentences with unusual syntax.[199] Additionally, different areas of law and different types of legal documents have their own terminology, standardized expressions, structure and hence their own distinctive word distributions. Criminal cases, for example, differ from consumer protection statutes, and tax statutes differ from lease agreements. In principle, it is reasonable to assume that legal text has different word distributions than non-legal text and that there are significant differences between different types of legal text. Consequently, many legal tasks may involve word distributions that were absent or uncommon in training corpora. Given the differences between legal and non-legal language, it can be assumed that LLMs trained on non-legal language will perform worse when deployed in tasks involving legal language.[200] Much will depend on the type of legal texts and, for example, whether the text in question was deliberately drafted to be more comprehensible to non-lawyers or whether it was directed at a more technical, expert audience. Ideally, LLMs should be able to work across different domains characterized by different lexical characteristics or refuse to generate an answer when provided with text or tasks it was not trained on.[201] At present, however, OOD robustness remains a problem[202] and is often regarded as an additional source of hallucinations.[203]

---

[195] Inbal Magar, Roy Schwartz, *Data Contamination: From Memorization to Exploitation*, PROC. 60th ANN. MEETING ASSOC'N COMPUTATIONAL LINGUISTICS 157 (2022).

[196] Bender et al., *supra* note 42, at 613

[197] When presented with OOD demonstrations containing different domains, the accuracy of GPT-4 deteriorated when presented with domains that were remote from the target domain, Boxin Wang et al, *supra* note X, at X; Brown et al., *supra* note 28, at 3.

[198] In the context of LLMs, OOD generalization is regarded as part of OOD robustness, which refers to their stability and performance when faced with various input conditions. This includes their ability to effectively handle diverse inputs, noise, interference, adversarial attacks, and changes in data distribution, among other factors, see: Lichao Sun et al., *supra* note 149, at 50-52.

[199] R. Friedrich, *Complexity and entropy in legal language*, 9 FRONT. PHYS. 67188 (2021); D. M. Katz & M.J. Bommarito, *Measuring the Complexity of the Law: the United States Code*, 22 ARTIF INTELL L 337 (2014); *See* DAVID MELLINKOFF, THE LANGUAGE OF THE LAW chs. 2–3 (1963).

[200] Kolt, *supra* note 145, at 94.

### D. Adaptation to downstream tasks

All legal tasks constitute downstream tasks. Lawyers do not, after all, specialize in text generation based on stochastic word prediction. The problems accompanying the adaptation of LLMs to legal tasks are both technical and legal in nature. Unfortunately, many techniques that seek to alleviate the problem of hallucinations or to adopt a language model to specialized downstream legal tasks may contribute to their existence.

### 1.  Fine-tuning

In the second phase of training, commonly referred to as fine-tuning, LLMs are adapted to specialized, domain-specific tasks. It is often overlooked that certain aspects of fine-tuning may become a source of hallucinations or, at least, contribute to their existence. In contrast to pre-training, which relies on unlabeled raw text, all fine-tuning techniques require some form of human input and thus expose the model to the possibility of introducing human error and bias.[204] In contrast to pre-training, which is purely statistical in nature and therefore unconcerned with such concepts as truthfulness or accuracy, many aspects of fine tuning require demonstrations of the desired output or evaluations of the generated output. Consequently, they require a reference to a legal ground truth or at least a reference to someone's *opinion* what the model's output should be.

Fine-tuning is traditionally associated with supervised machine learning, which involves training models on labelled input-output examples.[205] The main challenge, especially in the legal context, concerns the process of data labeling, or annotation. The process involves two groups of persons: those who annotate[206] and those who prepare annotation guidelines,[207] documents providing detailed instructions how to label individual text examples, including demonstrations of ideal

---

input-output pairs. Logically, the quality of the annotation depends on the quality of the guidelines and hence on the legal expertise of their creator(s). The problem is not, however, limited to the expertise of the creators but concerns the broader difficulty of formalizing legal rules in a given area, that is, interpreting or reconciling multiple legal sources in order to provide an unequivocal statement of the relevant law that can be translated into unambiguous annotation instructions. Inevitably, the latter will reflect the *opinions* of their creator(s), not an objective legal truth.

Interestingly, in supervised machine learning it is common to speak of "gold labels," examples considered to be correct. Gold labels assume the existence of a "ground truth," the perfect output or "the true answer we are asking our model to predict."[208] As indicated, such "legal ground truth" may not exist or be a question of subjective opinion.[209] As a result, the preparation of adequate labelling guidelines largely depends on the ability to reduce legal principles in a given area to a set of binary statements – and on the very possibility of ensuring that such statements represent the "ground truth." Logically, the quality of the labelling will also depend on the correct implementation of the guidelines by those who perform the actual annotation. To date, the creation of legal datasets has proven to be extremely resource intensive as it required the involvement of law students who had to be trained and supervised by experienced senior lawyers or academics.[210]

---

[208] Russel & Norvig, *supra* note 18, at 671.

[209] Reuben Binns, *Analogies and Disanalogies between Machine-Driven and Human-Driven Legal Judgement*, 1 J. CROSS-DISCIPLINARY RES. COMPUTATIONAL L. 1, 7–8 (2021) (observing that many legal questions do not have single correct answers, rendering it difficult to establish a "ground truth" to train machine learning models to perform legal tasks). D Litowitz, *Dworkin and Critical Legal Studies on Right Answers and Conceptual Holism*, (1994) 18(2) LEGAL STUDIES FORUM 135; KN LLEWELLYN, JURISPRUDENCE: REALISM IN THEORY AND PRACTICE (2011).

[210] Given the resources required to annotate legal documents, legal datasets are scarce and usually limited to relatively narrow tasks. For example, the Contract Understanding Atticus Dataset ("CUAD"), a dataset for legal contract review, involved a year-long effort by dozens of law student annotators, lawyers, and machine learning researchers. Annotators attended 70-100 hours of contract review sessions lead by experienced lawyers and were required to follow more than 100 pages of detailed annotation guidelines. Each annotation was verified by three additional annotators to ensure consistency and correctness. A conservative estimate of the monetary value of CUAD of is over $2 million as each of the 9283 pages were reviewed at least 4 times, each page requiring 5-10 minutes, assuming a rate of $500 per hour; see: Dan Hendrycks, Collin Burns, Anya Chen & Spencer Ball, *CUAD: An Expert-Annotated NLP Dataset for Legal Contract Review*, 35TH CONF. NEURAL INFO. PROCESSING SYS. DATASETS AND BENCHMARKS TRACK at 3–8 (2021); Similarly, the preparation of the Merger Agreement Understanding Dataset ("MAUD"), a legal reading comprehension dataset, required over 10,000 hours by law students and lawyers; See: Steven Wang et., *MAUD: An Expert-Annotated Legal NLP Dataset for Merger Agreement Understanding,* PROC. CONF. EMPIRICAL METH. NATURAL LANGUAGE PROCESSING 16369 (2023). CLAUDETTE involved the creation of a dataset for the automated identification and classification of unfair clauses in Terms-of-Service documents. Although the dataset was constructed from a corpus of "only" 50 online consumer contracts and involved a seemingly simply binary classification task, it required 18 months of analysis by legal experts and the creation of detailed labelling guidelines; see: Marco Lippi et al., *CLAUDETTE: An Automated Detector of Potentially Unfair Clauses in Online Terms of Service*, 27 AI & L. 117, 130–34 (2019); see also Ilias Chalkidis, Abhik Jana, Dirk Hartung, Michael Bommarito, Ion Androutsopoulos, Daniel Martin Katz & Nikolaos Aletras, *LexGLUE: A Benchmark Dataset for Legal Language Understanding in*

The above observations also apply to a recent technique, Reinforcement Learning from Human Feedback ("RLHF"), which involves a more complex suite of technologies but, nonetheless, relies on the quality of human input. In principle, RLHF aims to align LLMs with certain pre-set criteria by teaching them how to mimic certain outputs that were preferred by humans.[211] RLHF relies on demonstrations of the desired behavior, which are usually provided by labelers or obtained from text prompts submitted by LLM users.[212] These demonstrations are roughly comparable to the input-output examples encountered in supervised learning. Next, RLHF requires human evaluations of the model's outputs. The preferred outputs are used as training signals for a reward model,[213] which aims to replicate such outputs. It must be emphasized that RLHF leverages human preferences and aligns the model with the *opinions* of a specific group of people - not with any objective ground-truth. The quality of RLHF depends on the quality of the instructions given to the individual labelers,[214] who create the demonstrations and evaluate subsequent model outputs, as well as on their adherence to such instructions. [215] The instructions may, as in the case of annotation guidelines, inadequately represent the legal rules or, at least, present only one of many possible interpretations of legal principles. The labelers, usually crowd workers, may ignore the instructions altogether, and rely on their personal knowledge. [216] To aggravate matter, as labelers are often compensated per example, they may be disincentivized to read the detailed instructions that would assist them in providing optimal feedback.[217]  Financially motivated haste will generally trump careful deliberations regarding the quality of the generated output.

Another intrinsic limitation of this technique concerns the impossibility to evaluate the model's performance on difficult tasks[218] as it may often be extremely

---

*English,* ARXIV (Sept. 3, 2021), https: / /arxiv.org /pdf /2104.07782.pdf (surveying recent work on fine-tuning language models in the legal domain).

[211] One commonly used criterion today is "3H", which denotes helpful, honest, and harmless; *see:* Long Ouyang et al., *Training Language Models to Follow Instructions with Human Feedback*, 36 PROC. CONF. ON NEURAL INFO. PROCESSING SYS. (2022).

[212] *Id.*

[213] The reward model constitutes a numerical representation of a human preference, the goal the model seeks to achieve. The reward model predicts if a given output is good (high reward) or bad (low reward); see: Ouyang et al., *supra* note 211.

[214] An example of labeling instructions used by OpenAI: https://docs.google.com/document/d/1viWm6I2hBPFL2zqflj4s2it32FRbkETZpUS3CcVdFvo/edit?pli=1#heading=h.21o5xkowgmpj.

[215] Stephen Casper et al, *Open Problems and Fundamental Limitations of Reinforcement Learning from Human Feedback*, ARXIV (Sep. 11, 2023) https://arxiv.org/pdf/2307.15217.pdf.

[216] Nouha Dziri, Sivan Milton, Mo Yu, Osmar Zaiane & Siva Reddy, *On the Origin of Hallucinations in Conversational Models: Is it the Datasets or the Models?* PROC. 2022 CONF. NORTH AMERICAN CH. ASSOC'N. COMPUTATIONAL LINGUISTICS: HUMAN LANGUAGE TECH. 5271–5285 (2022)

[217] Niamh Rowe, *Millions of Workers Are Training AI Models for Pennies*, WIRED (Oct. 16, 2023) https://www.wired.com/story/millions-of-workers-are-training-ai-models-for-pennies/

[218] Stephen Casper et al, *Open Problems and Fundamental Limitations of Reinforcement Learning from Human Feedback*,  ARXIV (Sep. 11, 2023) https://arxiv.org/pdf/2307.15217.pdf.

difficult to provide a demonstration of an optimal output. The harder the task, the more complexity involved in evaluating the output and the greater the risk of incorrect feedback. Tasks requiring expert legal knowledge are particularly prone to succumb to subjective or even incorrect interpretations of the law. The entire process of RLHF is rife with opportunities to misrepresent or misinterpret the law or, at least, to side-step labelling instructions. RLHF may thus become a source of hallucinations or, at least, contribute to their existence. Again, abstracting from the practical problems concerning the preparation and implementation of the labelling guidelines, the broader problem concerns the difficulty of formalizing legal principles and stating the legal ground truth – the point of reference a particular output should be evaluated against.

### 2. Retrieval Augmented Generation

Another set of techniques aimed to adapt LLMs to specific tasks and to reduce their propensity to hallucinate is known as Retrieval Augmented Generation ("RAG").[219] To recall, LLMs lack access to external sources of information[220] and cannot verify facts, draw on existing knowledge or retrieve dynamic information such as weather reports or stock prices.[221] Their parametric knowledge not only abounds in incorrect information but becomes, as a matter of principle, fixed at a certain point in time. In principle, RAG involves a variety of techniques of retrieving relevant information from an external source and providing such information to the LLM.[222] In technical terms, it combines the model's parametric knowledge with external, non-parametric knowledge. [223] Despite its potential to improve the quality of the generated output,[224] RAG may also indirectly contribute to the problem of hallucinations. Logically, the quality of the model's output will largely depend on the source or, more pertinently, on the quality of the retrieved information or knowledge.[225] If the knowledge or information retrieved from an external source is incorrect, then the generated output will be incorrect.

---

[219] Patrick Lewis, et al., *Retrieval-augmented generation for knowledge-intensive NLP tasks*, ADVANCES IN NEURAL INF PROCESSING SYST 33 (2020), 9459–9474; Kelvin Guu, et al.,. *Retrieval augmented language model pre-training*, INT'L CONF ON MACHINE LEARNING 3929 (2020); G Mialon at al., *Augmented Language Models: a Survey*, ARXIV (Feb. 21, 2023) https://arxiv.org/pdf/2302.07842.pdf; Peng B, Galley M, He P, Cheng H, Xie Y, Hu Y, et al. *Check Your Facts and Try Again: Improving Large Language Models with External Knowledge and Automated Feedback*, ARXIV (Feb 28, 2023) https://arxiv.org/pdf/2302.12813.pdf.
[220] Ehud Karpas et al., *MRKL Systems: A modular, neuro-symbolic architecture that combines large language models, external knowledge sources and discrete reasoning*, ARXIV (May. 1, 2022) https://arxiv.org/abs/2205.00445.pdf.
[221] GPT-4 lacks knowledge of events that have occurred after April 2023, when its pre-training was completed; for an overview of various approaches to injecting knowledge into LLMs, see: Pedro Colon-Hernandez et al., *Combining pre-trained language models and structured knowledge* (Jul. 14, 2021) https://www.media.mit.edu/posts/combining-pre-trained-language-models-and-structured-knowledge/
[222] Some approaches involve specialized retrieval models, others connect the LLM to a search engine or add the retrieved information into the prompt, *see:* Lewis et al., *supra* note X, at X.
[223] Gautier Izacard et al., *ATLAS: Few-shot learning with retrieval augmented language models*, 24 JMLR 1-43 (2023)
[224] Lichao Sun et al., *supra* note 151, at 52.

Speaking of legal sources, the most logical option would be to augment LLMs with content from such popular legal databases like LexisNexis or Westlaw.It may still, however, be difficult to ensure that the information contained in such sources is not only correct but also presented in a manner that can be efficiently utilized by the LLM.  After all, LexisNexis and Westlaw provide structured databases of legal sources – not databases of "perfect" legal answers or ready-made solutions.  Ideally, LLMs should be connected to and augmented with legal knowledge bases – sources of legal *knowledge*, distilled from multiple legal primary and secondary sources and structured in a manner ensuring optimal "searchability."  The challenges of creating such knowledge bases are reminiscent of the problems of formalizing  knowledge, commonly referred to as the "knowledge acquisition and representation bottleneck."[226] As demonstrated in the context of legal expert systems,[227] which aimed to automate legal question answering by combining an inference engine with a knowledge base, the formalization of legal knowledge has proven extremely difficult given the need to identify, interpret and reconcile a large number of legal sources to extract the legal rules in a specific area as well as to represent such rules as finite sets of unambiguous if-then statements.[228] In a perfect scenario, the LLM could obtain "pre-digested" chunks of legal knowledge that could be used to generate substantively correct (or at least legally feasible) text.

In addition to the difficulties concerning the sources of legal information, additional challenges may derive from the retrieval methods, particularly with regards to the model's ability to select the most relevant documents or chunks of words from a given source. In other words, even a high quality  source of legal knowledge does not guarantee that the LLM will be able to utilize such knowledge!  As LLMs are often unable to distinguish between relevant and irrelevant information or to efficiently utilise multiple passages provided within their context, they may generate incorrect output even if provided with correct information directly within the prompt.[229] This has been vividly illustrated in recent studies where models were provided with the most relevant legal sources or even full answers to improve the accuracy of their output but were often unable to utilize such information.[230] At the

---

present state-of-the art LLMs may simply ignore the correct non-parametric knowledge and continue rely on their incorrect or outdated parametric knowledge. Paradoxically, in some instances RAG may even degrade the performance of LLMs as a result of "knowledge conflicts" between the retrieved knowledge and the model's existing parametric knowledge.[231]

### 3. Prompting

The quality of the generated output largely depends on the quality of the prompt, the instructions provided by the user of an LLM. Technical publications abound with novel prompting techniques that leverage the latent capabilities of LLMs and improve their performance in an increasing number of downstream tasks exceeding their original language generation objective.[232] Admittedly, carefully crafted prompts containing clear task descriptions and examples of reasoning chains, a skill known as prompt engineering,[233] could guide LLMs towards improved outputs and decrease their propensity to hallucinate. This assumption must, however, be qualified. While LLMs prompted with detailed instructions and informative demonstrations of the desired output, will generate better output than those provided with incorrect instructions and demonstrations, it must be acknowledged that in many instances users will lack the relevant technical and legal expertise to create optimal prompts. The process of prompting not only confirms that models understand neither the text of the prompt nor the output they generate,[234] but also highlights the importance of legal knowledge – or domain expertise in general – on the side of the users. Users must know *what* to ask and *how* to ask. Consequently, while hallucinations are largely attributable to the low-quality data used in training LLMs, user incompetence can also be a contributing factor. Two additional observations must be kept in mind:

First, LLMs are known to be extremely "prompt-sensitive" and will generate dramatically different outputs depending on minimal variations in the wording and structure of semantically equivalent prompts.[235] Even minuscule differences in the

---

Savelka et al., *Explaining Legal Concepts with Augmented Large Language Models (GPT-4)*, ARXIV (Jun. 22, 2023) https://arxiv.org/pdf/2306.09525.pdf.

[231] Alex Mallen et al., *When Not to Trust Language Models: Investigating Effectiveness of Parametric and Non-Parametric Memories*, PROC. 61st ANNUAL MEETING ASSOC. FOR COMPUTATIONAL LINGUISTICS 9802–9822 (2023); Cheng Qian et al., *Merge Conflicts! Exploring the Impacts of External Distractors to Knowledge Graphs,* ARXIV (Sep. 15, 2023) https://arxiv.org/pdf/2309.08594.pdf.

[232] Wei et al., *Chain-of-Thought Prompting: Eliciting Reasoning in Large Language Models* (2022) https://arxiv.org/abs/2201.11903; Takeshi Kojima et al., *Large Language Models Are Zero-Shot Reasoners*, 36 CONF. NEURAL INFO. PROCESSING SYS. (2022), https://arxiv.org/pdf/2205.11916.pdf; Shunyu Yao et al., *Tree of Thoughts: Deliberate Problem Solving with Large Language Models,* ARXIV (Dec. 3, 2023) https://arxiv.org/pdf/2305.10601.pdf.

[233] Fangyi Yu, Lee Quartey, and Frank Schilder, *Legal prompting: Teaching a language model to think like a lawyer*, ARXIV (Dec. 8, 2022) https://arxiv.org/pdf/2212.01326.pdf.

[234] Albert Webson & Ellie Pavlick, *Do prompt-based models really understand the meaning of their prompts?* PROC. 2022 CONF. NORTH AMERICAN CHAP. ASSOC. COMPUTATIONAL LINGUISTICS: HUMAN LANGUAGE TECHNOLOGIES, 2300 (2022).

[235] J.D. Zamfirescu*: How Non-AI Experts Try (and Fail) to Design LLM Prompts*, PROC. 2023 CHI CONF. HUMAN FACTORS COMPUTING SYSTEMS (2023); Adam Roegiest et al., *Questions about contracts: Prompt Templates for Structured Answer Generation,* PROC. NATURAL LANGUAGE

choice or arrangement of words may lead the model to produce different answers to a legal question. The word distributions in the prompt seem to be more important than the substance of the question, not to mention the correctness of the generated answer. For example, a study examining whether LLMs can understand consumer contracts, established that GPT-3 was attuned to the wording of the prompt that contained a question about a contractual provision but largely indifferent to language variations in the provision itself.[236] In other words, the text of the prompt was more important than the text the model was instructed to examine! Moreover, LLMs cannot differentiate between relevant and irrelevant input, between the text that should affect the answer to a legal question and the text that bears no relevance to such question. It has been demonstrated that even the most advanced LLMs change their answer to simple factual questions when their prompt contains irrelevant sentences.[237] Seemingly, the latter "inadvertently upweight the token probability of incorrect answers by virtue of those tokens appearing in the context."[238]

Second, LLMs are technically incapable of correcting the user's prompt, not to mention questioning user instructions. They are, in other words, biased "towards accepting legal premises that are not anchored in reality and answering queries accordingly." [239] Without the necessary legal expertise, users may prompt the model with text containing incorrect opinions or assumptions that will guide the model towards incorrect answers. A user could, for example, prompt the model to explain the concept of offer and acceptance in the context of criminal law or inquire about the legal requirements of granting patents to GenerativeAI. Unfortunately, LLMs have been shown to be sycophants that tended to agree with the opinions or assumptions contained in a prompt and generated text accommodating such opinions or assumptions,[240] even if the latter contained incorrect premises or assumptions.[241] Consequently, even when provided with nonsensical instructions ("electron for president!") or false premises ("whenever it rains, pigs fly!"), LLMs will still generate coherent and superficially plausible answers.

---

PROCESSING WORKSHOP 62 (2023); Frieda Rong, *Extrapolating to Unnatural Language Processing with GPT-3's In-context Learning: The Good, the Bad, and the Mysterious* (May 28, 2020) http://ai.stanford.edu/blog/in-context-learning/.

[236] Kolt, *supra* note 145, at 118 ("[G]iven that performance is so sensitive to the wording of the questions, it is somewhat puzzling that performance is altogether insensitive to the language of the contracts themselves. One possible explanation is that GPT-3, like other language models, operates by predicting the next word in a sequence. The question (not the contract) is the final part of the prompt and, therefore, has an outsized impact on the model's predictions.")

[237] Freda Shi et al., *Large language models can be easily distracted by irrelevant context*, INT'L. CONF. MACHINE LEARNING 31210–31227 (2023).

[238] Jason Weston and Sainbayar Sukhbaatar, *System 2 Attention (is something you might need too)* ARXIV (Nov. 20, 2023) https://arxiv.org/pdf/2311.11829.pdf.

[239] Dahl et al., *supra* note 8, at 12.

[240] Ethan Perez et al. *Discovering language model behaviours with model-written evaluations*, FINDINGS ASSOC COMPUTATIONAL LINGUISTICS 13387 (2023). https://aclanthology.org/2023.findings-acl.847/. Mrinank Sharma et al. *Towards understanding sycophancy in language models*, ARXIV (Oct. 27, 2023) https://arxiv.org/pdf/2310.13548.pdf.

[241] Dahl et al., *supra* note 8, at 12.

# V. Benchmarks and Bar Exams

It is necessary to address two elephants in the room, two anticipated objections to all of the aforementioned shortcomings of LLMs. After all, despite such shortcomings, LLMs not only generate impressive, human-like text but also excel at a wide range of benchmarks testing their ability to reason, understand and to exhibit common sense. More importantly, GPT-4 has also passed the Bar exam – this feat alone should dispel any doubts regarding the potential of LLMs in general!

## A. Benchmarks

The actual capabilities and limitations of LLMs are difficult to estimate given that both technical and legal literature abound in proclamations that LLMs can, for example, "answer factual questions,"[242] "reason"[243] or "understand."[244] Such proclamations must, however, be approached with caution as they generally refer to benchmark performance, that is, the performance of models on artificially created and narrowly defined tasks that generally bear little resemblance to the broader skills they purport to measure. In practice, stellar benchmark performance does not mean that a model can perform in real-world scenarios[245] or that it has developed the ability to reason or understand.[246] In fact, NLP research is often criticized for overly focusing on achieving high scores on benchmarks that test a specific instance of a skill, instead of developing LLMs that actually have the general skill.[247] Moreover, as the tasks comprised in benchmarks are known in advance, models can be developed with the benchmark(s) in mind,[248] not with the broader purpose of developing the *actual* ability purportedly measured thereby.[249] Two observations are apposite.

First, although LLMs have performed surprisingly well on benchmarks testing language understanding[250] or common-sense reasoning[251] they often relied on heuristics, or "technical shortcuts,"[252] and exploited certain biases in those datasets.[253] Even the notoriously hard Winograd Schema Challenge, which involves the complex task of pronoun disambiguation and was specifically designed to avoid such heuristics,[254] permitted models to perform well despite not actually understanding the relevant text.[255] Once a model was provided with adversarial examples, i.e., inputs designed to fool the model and test its actual understanding of the task, its performance often fell back to chance.[256] The ability to exploit 'technical shortcuts" is further aggravated by the fact that as models are trained on content sourced from the internet, such content may contain – or be contaminated by - the same data that is subsequently used in evaluating their performance.[257] Such contamination resembles disclosing the exam questions to students before the exam and, unsurprisingly, leads to unrealistic estimates of model performance. Models can simply memorize solutions in their training sets instead of learning to generalize to answer new questions.

---

[242] Aitor Lewkowycz, et al., *Solving quantitative reasoning problems with language models*, ARXIV (Jul. 1, 2022) https://arxiv.org/pdf/2206.14858.pdf.

[243] Takeshi Kojima et al., *Large Language Models Are Zero-Shot Reasoners,* 36 CONF. NEURAL INFO. PROCESSING SYS. (May 24, 2022), https://arxiv.org/pdf/2205.11916.pdf.

[244] Refer back

[245] Francois Chollet, *On the Measure of Intelligence,* ARXIV (Nov. 5, 2019). https://arxiv.org/pdf/1911.01547.pdf.

[246] Inioluwa Deborah Raji et al., *AI and the Everything in the Whole Wide World Benchmark*, 35 PROC. CONF. ON NEURAL INFO. PROCESSING SYS. (NeurIPS 2021).

[247] Melanie Mitchell, *On Crashing the Barrier of Meaning in Artificial Intelligence*, 41 AI MAGAZINE 86-92 (2021) https://doi.org/10.1609/aimag.v41i2.5259.

[248] Chollet, *supra* note 245, at 18.

[249] John Nay, *Law Informs Code: A Legal Informatics Approach to Aligning Artificial Intelligence with Humans*, 20 NORTHWESTERN J. TECH. & I. P. 2, 31 ((2022) referring to "Goodhart's Law," "when a measure becomes a target, it ceases to be a good measure."

[250] P. Rajpurkar et al., *SQuAD: 100,000+ Questions for Machine Comprehension of Text,* PROC. 2016 CONF. EMPIRICAL METHODS IN NATURAL LANGUAGE PROCESSING 2383 (2016).

[251] R. Zellers et al., *SWAG: A Large-Scale Adversarial Dataset for Grounded Commonsense Inference,* PROC. 2018 CONF. EMPIRICAL METHODS IN NATURAL LANGUAGE PROCESSING 93 (2018).

Second, before praising a model's performance on a given benchmark, it is worth investigating whether the benchmark adequately represents the relevant skill, that is, whether the tasks comprised therein can be regarded as proxies for the skill in question.[258] For example, the aforementioned Winograd Schema Challenge is supposed to test common sense with a series of complex pronoun disambiguation tasks. It is increasingly recognized, however, that pronoun disambiguation may represent only a small subset of the commonsense knowledge required to understand and use natural language.[259] Despite its complexity, it is not necessarily indicative of common sense in general. Similarly, although the popular benchmarks GLUE[260] and SuperGLUE[261] are supposed to enable "a general-purpose evaluation of language understanding,"[262] a closer examination reveals that although the said benchmarks purport to measure linguistic competence together with common-sense reasoning and world knowledge, their constituent tasks are arbitrary and do not necessarily represent those skills. [263] It has even been observed that at present, none of the existing benchmarks covers a sufficient number of *diverse* tasks that would represent "human linguistic activity"[264] or "general understanding."[265] Arguably, such benchmark would have to include hundreds if not thousands of discrete tasks that *taken together* could be regarded as representative of the model's ability to reason or understand. It can, however, be questioned whether such "list of representative tasks" could exist in the first place.

---

[252] Bender et al., *supra* note 42, at 616

[253] Timothy Niven and Hung-Yu Kao. *Probing neural network comprehension of natural language arguments*, PROC. 57th ANN. MEETING ASSOC. COMPUTATIONAL LINGUISTICS (2019) pages 4658–4664, https://www.aclweb.org/anthology/P19-1459.

[254] Hector J. Levesque et al., *The Winograd Schema Challenge*, LOGICAL FORMALIZATIONS OF COMMONSENSE REASONING: PAPERS FROM THE 2011 ASSOCIATION FOR THE ADVANCEMENT OF ARTIFICIAL INTELLIGENCE.

[255] K Sakaguchi et al.,. *WinoGrande: An Adversarial Winograd Schema Challenge at Scale,* 64 COMMUNICATIONS of the ACM (2019) https://dl.acm.org/doi/10.1145/3474381

[256] Vid Kocijan et al., *The defeat of the Winograd Schema Challenge,* ARTIF. INTEL. 103971 (2023), (discussion of shortcomings of WSC)

[257] I Magar and R. Schwartz, *Data Contamination: From Memorization To Exploitation*, PROC. 60th ANN MEETING ASSOC. COMPUTATIONAL LINGUISTICS 157 (2022).

[258] The question whether a particular evaluation accurately represents and measures the construct is referred to as 'construct validity," *see;* Kapoor et al., *supra* note 8, at 5.

[259] Kocijan et al., *supra* note X,  at X.

[260] Alex Wang et al., *GLUE: A Multi-Task Benchmark and Analysis Platform for Natural Language Understanding*, 7th INT'L CONF ON LEARNING REPRESENTATIONS (2019).

[261] Alex Wang et al., *SuperGLUE: A stickier benchmark for general-purpose language understanding systems*, 33 CONF. NEURAL INFO. PROCESSING SYS. 3266 (December 2019) https://dl.acm.org/doi/10.5555/3454287.3454581.

[262] *Id.* 2

This point seems to be indirectly confirmed by some benchmarks in the legal domain.[266] Of particular interest is LegalBench, a collaborative benchmark examining whether LLMs can execute tasks that require legal reasoning.[267] Acknowledging the different types of legal tasks, the benchmark comprises an online repository allowing domain experts to submit tasks that evaluate, or represent, different forms of legal reasoning, ranging from Issue, Rule, Application and Conclusion (IRAC) tasks, which represent standard legal reasoning, to more general-purpose tasks, such as clause classification and information extraction.[268] While the significance of this unprecedented domain-specific effort must not be downplayed, the very existence of LegalBench illustrates the difficulty of creating a comprehensive list of tasks that measure the ability to engage in legal reasoning. The problem is not that the individual tasks are extremely narrow, often confined to a binary classification of one contractual clause,[269] but that there are so many tasks in the first place!

Is it worthwhile measuring model performance with a multitude of discrete tasks that exemplify specific instances of legal reasoning  or is it more advisable to focus on the broader tasks of causal or deductive reasoning? Is it worthwhile investigating whether LLMs can identify whether a *clause* describes how user information is protected or is it better to establish whether LLMs that can reason by analogy? It is tempting to discard many of the benchmarking efforts with the observation that if LLMs struggle with the basic building blocks of reasoning, such as causality and common sense, testing specific instances of reasoning may seem like an academic exercise.  Nonetheless, as the evaluation of legal skills is inherently difficult, it is useful to have *some* clear measures of performance.[270] Such measures must not, however, be blindly relied on…

---

[263] Raji et al., *supra* note 246,  at 3.

[264] *Id.*  at 7.

[265] *Id.* at 7.

[266] Legal benchmarks: Ilias Chalkidis et al., *LexGLUE: A Benchmark Dataset for Legal Language Understanding in English* PROC. 60th ANN MEETING ASSOC. COMPUTATIONAL LINGUISTICS 4310 (2022), which contains multiple datasets and a variety of tasks; J Niklaus et al., *Lextreme: A multi-lingual and multi-task benchmark for the legal domain*, FINDINGS ASSOC. COMPUTATIONAL LINGUISTICS 3016 EMNLP 2023.

[267] Neel Guha, Daniel E. Ho, Julian Nyarko, Christopher Ré, *LegalBench: Prototyping a Collaborative Benchmark for Legal Reasoning*, ARXIV (Sep. 13, 2022) https://arxiv.org/pdf/2209.06120.pdf.

[268] At the time of writing LegalBench contains 162 tasks, see: https://github.com/HazyResearch/legalbench.

[269] For example, the task of identifying "if the clause provides that all Confidential Information shall be expressly identified by the Disclosing Party," which represents one of the skills of legal interpretation, *see*: https://github.com/HazyResearch/legalbench/tree/main/tasks/contract_nli_explicit_identificatin

## B. ChatGPT passes the Bar Exam

Unlike artificial benchmarks, the Bar Exam is a test designed to measure the actual human ability to solve legal problems. Unsurprisingly, the fact that GPT-4 passed this exam has raised significant concerns in the legal profession, not to mention predictions of imminent legal unemployment.[271] This "achievement" must, however, be approached with considerable caution. To date, GPT-4 has passed the exam once and the experiment has not been replicated. It must be a remembered that GPT-4 is a commercial product, created by a for-profit entity with an incentive to downplay the model's limitations and to overemphasize its achievements. To evaluate the significance of GPT-4's performance on the Bar Exam, one would require insights into the model's training method and training data.[272] Citing competitive pressures, however, OpenAI provided no details about the construction of the dataset and the training methods.[273] In particular, no assurances were provided that the model did not see any of the data it was tested on during training. As a matter of principle, a model should only be tested on new data as the presence  of test data in the training set would provide an overly optimistic evaluation of the model's capabilities.[274] Although OpenAI asserted that it did not do any specific training for the Bar Exam, it must be assumed that during training the model has potentially seen all prior Bar Exam questions as well as their correct responses.[275] As indicated, the text corpora used in pre-training are often "contaminated" with downstream test sets.[276] When evaluating GPT-4 on *traditional benchmarks*, OpenAI emphasized that it "ran contamination checks for test data appearing in the training set."[277] The Bar Exam, however, is *not* a traditional benchmark and the relevant documentation is silent whether such contamination checks were performed.

Technicalities aside, passing the Bar Exam (a single time) provides limited insights as to how the model would perform when confronted with "real life" legal problems. Performing well on one version of a problem does not mean that the model understands the problem or that it will be able to solve similar problems in the future.[278] The Bar Exam is known to "overemphasize subject-matter knowledge and underemphasize real-world skills."[279] At the same time, such "real-world skills" are also more difficult to measure in an standardized and computer-administered manner.

---

[270] Kocijan et al., *supra* note 256, at 11.

[271] Daniel Martin Katz, Michael James Bommarito, Shang Gao and Pablo Arredondo, *GPT-4 Passes the Bar Exam* (March 15, 2023), SSRN: https://ssrn.com/abstract=4389233 or http://dx.doi.org/10.2139/ssrn.4389233

[272] For a detailed critique, *see*: *see*: Kapoor et al., *supra* note 7; Eric Martínez, *Re-Evaluating GPT-4's Bar Exam Performance*, May 2023. URL https://papers.ssrn.com/abstract= 4441311.

[273] GPT-4 Technical Report, *supra* note 16, at 2.

[274] ("It is possible that evaluations such as OpenAI's claims about bar exam performance are overoptimistic due to contamination, but it is hard to know for sure due to the training and fine tuning data being proprietary.") *see*: Kapoor et al., *supra* note 7, at 4.

[275] Bubeck et al., *supra* note 14, at 7.

[276] I Magar and R. Schwartz, *supra* note 257.

[277] GPT-4 Technical Report, *supra* note 16, at 6.

[278] https://aiguide.substack.com/p/did-chatgpt-really-pass-graduate M Mitchell

Notwithstanding the foregoing, GPT-4's performance is unquestionably impressive given that some components of the Bar Exam required the drafting of memos in the context of a complex legal scenarios, including domestic relations, criminal law, and legal ethics. Interestingly, GPT-4 demonstrated the largest improvement in the areas of contract law and evidence. It was not, however, required to draft or interpret a contract, (not to mention: question witnesses or gather evidence) but to answer multiple-choice questions about contract law and about evidence. Answering questions *about* legal principles, however, differs from applying these principles against the background of a specific transaction or in the context of a specific case. It would be interesting to observe GPT-4's performance on tasks involving the interpretation of statutes[280] and contracts, given that its predecessor, GPT-3, scored rather low.[281]

# VI. Possible Solutions?

It is important to address the question whether the current limitations of LLMs can be addressed with technology or mitigated by their users, the lawyers who prompt LLMs to obtain high-quality outputs.

## A. Scaling LLMs

It is often implied that once LLMs are provided with more training data and more computational power, they will 'scale up' and develop understanding, commonsense, and other abilities that will remedy their current shortcomings, including their propensity to hallucinate. After all, *large* language models have already displayed unanticipated, or emergent, abilities that "exceeded" their original training. Prominent examples include their ability to learn from a few examples provided in the prompt[282] or their proficiency at coding.[283] In the context of LLMs, emergence generally refers to abilities that are absent in smaller models but "appear" in larger ones and could not have been predicted by simply extrapolating from the performance of smaller models.[284]  While it cannot be doubted that model size has evident practical consequences,  there is not guarantee that scaled up models will naturally develop additional capabilities. To the contrary. There is growing consensus that such skills like understanding, common sense, and reasoning will *not* emerge from more powerful LLMs.[285] Scaling up cannot solve the grounding problem required for understanding or equip LLMs with a world model required for common sense and causal reasoning.

---

[279] Ben Bratman, *Improving the Performance of the Performance Test: The Key to Meaningful Bar Exam Reform*, 83 U MISSOURI-KANSAS CITY L. REV. 565, 568 (2015).

[280] GPT-3 performance on statutory reasoning was unimpressive, especially given that it had to reason about a simplified statutory dataset covering 9 sections of the U.S. tax code, see: Nils Holzenberger, Andrew Blair-Stanek & Benjamin Van Durme, *A Dataset for Statutory Reasoning in Tax Law Entailment and Question Answering*, PROC. 2020 NATURAL LEGAL LANGUAGE PROCESSING WORKSHOP at 4–5 (2020).

[281] GPT-3 could not understand or reason about consumer contracts, see: Kolt, *supra* note 145, at 71.

[282] Brown et al., *supra* note 30, at 6–7.

[283] Bubeck et al., *supra* note 14, at 31.

Scaling will not overcome the inherent limitations of transformer-based language models, which can *by definition* only excel at word prediction. It has, in fact, been demonstrated that many tasks requiring causal reasoning and general world knowledge exhibit so-called flat scaling curves, that is, that more powerful models do not perform those tasks above-random chance.[286] It has also been established that the quality of the generated output often decreases with increased model size,[287] that larger models are often less truthful than their smaller predecessors[288] and that the performance of transformer-based LLMs rapidly decays when confronted with tasks of increased complexity.[289] In principle, making models larger seems to make them more fluent but not necessarily more trustworthy or reliable.[290] Arguably, with  larger models hallucinations may not become less frequent but simply more difficult to detect. The size of the model does not, after all, change its inherent architecture or the manner it operates. Irrespective of their size, autoregressive transformers will continue to predict the next word based on the previous word and will not magically acquire the ability to "reach outside" their training corpus.

---

[284] Jason Wei et al., *Emergent Abilities of Large Language Models*, TRANSACTIONS ON MACH. LEARNING RSCH. (Aug. 2022), https://openreview.net/pdf?id=yzkSU5zdwD; some doubts have, however, been cast on the very concept of emergence. Seemingly, the phenomenon can be explained  by the choice of metric used to evaluate the generated outputs, challenging the theory that certain capabilities in LLMs derive from their scale, see: Rylan Schaeffer, Brando Miranda, Sanmi Koejo, *Are Emergent Abilities of Large Language Models a Mirage?* 37 PROC. CONF. ON NEURAL INFO. PROCESSING SYS. (2023); for detailed analysis of the scaling of language models, *see* Jared Kaplan et al, *Scaling Laws for Neural Language Models*, ARXIV (Jan. 23, 2020), https://arxiv.org/abs/2001.08361; Tom Henighan et al., *Scaling Laws for Autoregressive Generative Modeling*, ARXIV (Oct. 28, 2020), https://arxiv.org/abs/2010.14701.

[285] Anil Ananthaswamy, *In AI, is Bigger Better?* (2023) 615 NATURE 202, 204; Rae et al, *Scaling language models: Methods, analysis & insights from training Gopher,* ARXIV (Jan. 21, 2022) https://arxiv.org/pdf/2112.11446.pdf; Steven Levy, *How Not to Be Stupid About AI, With Yann LeCun,* WIRED (Dec. 22, 2023) https://www.wired.com/story/artificial-intelligence-meta-yann-lecun-interview/("Machine learning is great. But the idea that somehow we're going to just scale up the techniques that we have and get to human-level AI? No. We're missing something big to get machines to learn efficiently, like humans and animals do.")

[286] Mirac Suzgun et al. *Challenging BIG-Bench Tasks and Whether Chain-of-Thought Can Solve Them*, FINDINGS 2023 ASSOC'N. COMPUTATIONAL LINGUISTICS 13003 (2023).

[287] Antonio Valerio Miceli-Barone et al., *The Larger They Are, the Harder They Fail: Language Models do not Recognize Identifier Swaps in Python.* ARXIV (May. 24, 2023) https://arxiv.org/pdf/2305.15507.pdf.

[288] Stephen Lin et al., *TruthfulQA: Measuring How Models Mimic Human Falsehoods*, PROC OF THE 60th ANNUAL MEETING OF THE ASSOC FOR COMPUTATIONAL LINGUISTICS 3214 (2021) TruthfulQA aims to measure the model's ability to identify when a claim is true in the sense of "literal truth about the real world", and not in the context of a belief system or tradition. While strong performance on TruthfulQA does not imply that a model will be truthful in a specialized domain, poor performance indicates a lack of robustness.

[289] Nouha Dziri et al., *Faith and Fate: Limits of Transformers on Compositionality,* ARXIV (May 29, 2023) https://arxiv.org/pdf/2305.18654.pdf, the trend remains the same for few-shot prompting.

[290] Romal Thoppilan et al., *LaMDA: language models for dialogue applications,* ARXIV  (Feb. 10, 2022) https://arxiv.org/pdf/2201.08239.pdf.

It has also been established that increasing model size does not significantly improve parametric knowledge with regards to less popular words and hence potentially reduce the occurrence of hallucinations in specialized domains.[291] In their present form, autoregressive transformers will continue to memorize those strings of words that appear in their training corpora with greater frequency and, in principle, will not perform well on tasks that require domain-specific knowledge that is underrepresented or absent from such corpora.

It is also necessary to briefly address the assumption that language models are gradually becoming more trustworthy and reliable.[292] While it cannot be denied that the performance of some models shows consistent improvement with regards to *some* specific, narrowly defined tasks, the overall "behaviour" of closed models, such as GPT-4, is often unpredictable and not necessarily better. This is commonly attributed to ongoing updates and the introduction of stronger moderation mechanisms that aim to hinder the generation of harmful content.[293] Such changes need not necessarily result in a performance degradation but, in practice, make the model less predictable and hence reliable.[294] How can LLMs be integrated into a legal workflow if their outputs are not reproducible, that is, if models generate different responses to identical prompts?

## B. Availability of Training Data

The idea of making models larger is intrinsically related to the possibility of training them on larger amounts of data. It cannot, however, be assumed that the technical ability to train models on more data will be accompanied by the continued availability of such data.[295] There are concerns surrounding the inadvertent disclosure of private information contained in the training data,[296] as well as a growing realization that LLMs have been trained on content that is subject to copyright,[297] such as books and newspaper articles. It can be anticipated that both private persons and content creators will increasingly object to their data and their content being used to train LLMs and that, as a result, the amount of data available for training will gradually decrease. To aggravate matters, it is becoming apparent that an increasing amount of online content is generated by LLMs and used to train subsequent generations of models.[298] Training models on such generated text will not only further propagate falsehoods and misinformation[299] but, given its synthetic nature, result in a gradual degeneration of a model's underlying data distribution,[300] a phenomenon known as "model collapse." Consequently, LLMs trained on text generated by other models will have progressively lower quality parametric knowledge and also decline in their general predictive capabilities.

---

[291] Alex Mallen et al., *When Not to Trust Language Models: Investigating Effectiveness of Parametric and Non-Parametric Memories*, PROC 61ST ANN. MEET. ASSOC'N. COMPUTATIONAL LINGUISTICS 9802 (2023).

[292] For a sobering overview *see*: Boxin Wang et al, *DECODING TRUST: A Comprehensive Assessment of Trustworthiness in GPT Models,* ARXIV (Jan. 5, 2024), https://arxiv.org/pdf/2306.11698.pdf.

[293] Lingjiao Chen, Matei Zaharia, James Zou, *How is ChatGPT's behavior changing over time?* ARXIV (Oct. 31, 2023) https://arxiv.org/pdf/2307.09009.pdf.

[294] The inconsistent behavior of such models like GPT-3.5 or GPT-4 is commonly associated with so-called "performance drift." *Id.*

[295] *See* Tammy Xu, *We Could Run Out of Data To Train AI Programs*, MIT TECH. REV. (Nov. 24, 2022), https://www.technologyreview.com/2022/11/24/1063684/we-could-run-out-of-data-to-train-ai-language-programs. (less high-quality data will be available)

[296] Amy Winograd, *Loose-Lipped Large Language Models Spill Your Secrets: The Privacy Implications of Large Language Models*, 36 HARV. J. L. & TECH. 616, 623-628 (2023). (observes that private information memorized in the model's training data may be vulnerable to exposure as adversaries can attack LLMs to elicit such memorized information.)

## C. Quality of Training Data

Many of the shortcomings of LLMs derive from the low quality of the data they are pre-trained on. Bombarded with headlines proclaiming the inevitable progress in language modelling, it is easy to forget the classic saying in computer science "garbage in, garbage out," the fact that low quality input will inevitably produce low quality output.[301] The common fascination with the fact that language models can be pre-trained "directly on raw, messy, real-world data, without the need for carefully curated, and human labelled data sets"[302] deflects attention from the simple truth that such "raw, messy, real-world data" abounds with false information that becomes part of the models' parametric knowledge that will inevitably taint the text generated in various downstream tasks. It is difficult to imagine how models imbued with incorrect parametric knowledge could generate factually correct output or at least output that is legally feasible. Hallucinations are, after all, regarded as a direct consequence of incorrect parametric knowledge.[303] From a purely statistical perspective, a good first step leading to the reduction of legal hallucinations would require that the training corpora contained "high-probability sentences" representing correct statements of the law.[304]

---

[297] Michael M. Grynbaum and Ryan Mac, *The Times Sues OpenAI and Microsoft Over A.I. Use of Copyrighted Work,* N.Y. TIMES (Dec. 27, 2023) (The defendats should be held responsible for "billions of dollars in statutory and actual damages" related to the "unlawful copying and use of The Times's uniquely valuable works." Any chatbot models and training data that use copyrighted material should be destroyed.) For an overview of the Status of all U.S. copyright cases against AI companies as of Jan. 23, 2024, see: https://chatgptiseatingtheworld.com/2024/01/23/status-of-all-u-s-copyright-cases-v-ai-cos-jan-23-2024-including-mtd-decision-in-doe-1-v-github-by-judge-tigar/ ; Mark A. Lemley & Bryan Casey, *Fair Learning,* 99 TEX. L. REV. 743, 760–79 (2021) (exploring whether the doctrine of fair use permits machine learning models to be trained on copyrighted data); Kate Nibbs, *Meet the Lawyer Leading the Human Resistance Against AI,* WIRED (Nov. 22, 2023) https://www.wired.com/story/matthew-butterick-ai-copyright-lawsuits-openai-meta/

[298] Abeba Birhane, *Synthetic Data Is a Dangerous Teacher*, WIRED (Jan. 8, 2024) https://www.wired.com/story/synthetic-data-is-a-dangerous-teacher/ (Observing current trends, Birhane predicts that in 2024, "a very significant part of the training material for generative models will be synthetic data produced from generative models. Soon, we will be trapped in a recursive loop where we will be training AI models using only synthetic data produced by AI models. Most of this will be contaminated with stereotypes that will continue to amplify historical and societal inequities. Unfortunately, this will also be the data that we will use to train generative models applied to high-stake sectors including medicine, therapy, education, and law. […] By 2024, the generative AI explosion of content that we find so fascinating now will instead become a massive toxic dump that will come back to bite us.")

[299] Hans W. A. Hanley and Zakir Durumeric, *Machine-Made Media: Monitoring the Mobilization of Machine-Generated Articles on Misinformation and Mainstream News Websites,* ARXIV (Jan. 17, 2024) https://arxiv.org/pdf/2305.09820.pdf. (who found that among reliable/mainstream news websites, synthetic articles increased in prevalence by 55.4% (0.91% of news articles in January 2022 to 1.42% in May 2023) while among unreliable/misinformation websites, the prevalence increased by 457% (0.38% of news articles in January 2022 to 2.14% in May 2023). When examining the content of synthetic articles, the authors found that while mainstream/reliable news websites generally utilized synthetic articles to report on financial and business-related news, misinformation/unreliable news websites have reported on a wide range of topics ranging from world affairs to human health.)

[300] Ilia Shumailov et al., *The Curse Of Recursion: Training On Generated Data Makes Models Forget,* ARXIV (May, 31 2023), https://arxiv.org/pdf/2305.17493.pdf; This phenomenon differs from *catastrophic forgetting* as "models do not forget previously learned data, but rather start misinterpreting what they believe to be real, by reinforcing their own beliefs."

The problem of low-quality training data is increasingly recognized in technical scholarship, which emphasizes that performance gains can only be achieved once training data is evaluated as to its accuracy.[305] At the same time, the unprecedented progress in language modelling is largely attributable to the very ability to train models on large swaths of unlabeled raw data taken from the internet. Having to improve the quality of such data would not only drastically increase the costs of pre-training LLMs but also stand in conflict with the core premise of this training phase.

### D. Improved Human Interventions & Input

Most efforts to improve the quality of the training corpora will require some form of human involvement, be it in the form of filtering such data or carefully selecting the sources to be included in the training corpora. Such involvement may, however, inadvertently introduce the risk of bias and human error. Who would curate the text? Who would decide which online sources to include and which to exclude? Indirectly, the latter question translates into: who would decide which opinions or types of content deserve representation in the training corpora? Should models be trained on text from Wikipedia, Reddit, or X (formerly known as Twitter)? Mapping this question onto the legal context, should language models be trained on *all* papers uploaded to SSRN or only on selected primary sources such as statutes and case law? What is the greater evil: training models on "uncurated" text containing bias, misinformation, and falsehoods but representative of the opinions of a large cross-section of society or training them on text selected by a group of experts in a given area?  In the legal context, the problem is not confined to the proliferation of misinformation and bias but extends to the proliferation of uninformed opinions *about* the law or specific legal areas made by persons without legal training.

---

[301] R. Stuart Geiger et al., *"Garbage in, garbage out" revisited: What do machine learning application papers report about human-labeled training data?* 2 QUANTITATIVE SC. STUD. 795 (2021).

[302] MUSTAFA SULEYMAN & MICHAEL BHASKAR, THE COMING WAVE (Crown 2023, NY) 65

[303] McCoy et al., *supra* note 9, at 11, 31-32, Lichao Sun et al., *supra* note 151, at 26 ("This behavior of generating inaccurate information can be attributed to imperfect training data. Given that LLMs are trained on vast volumes of text collected from the internet, the training dataset could encompass erroneous details, obsolete facts, or even deliberate misinformation.").

[304] McCoy et al., *supra* note 9, at 31, 32, 47.

[305] Yue Zhang et al., *supra* note 65, at 10.

Leaving aside the cost implications of curating the enormous text corpora, it must be remembered that, at some stage, most forms of human intervention will not only confront the problem of selecting the "worthy sources of legal knowledge" but also the challenge of establishing the "ground truth(s)" in various legal areas. The latter challenge will be particularly visible in the context of fine-tuning, as the quality of the training data will largely depend on the ability to provide correct examples or demonstrations of optimal output. A mislabeled dataset or incorrect demonstrations and feedback will not only fail to adapt the model to downstream tasks [306] but, most likely, lead to more hallucinations – at least if one associates this term with legally incorrect statements or statements that diverge from established legal doctrine.

It can be theorized that RLHF could significantly improve the fine-tuning process and thus model performance if the human feedback was replaced with *expert* human feedback. Such expert feedback would have to involve experienced lawyers, scholars, or even judges who could provide the model with the most accurate demonstrations of the preferred output and subsequently evaluate the model's performance in order to refine the reward model. While the resulting technique, renamed as RLEHF (Reinforcement Learning from *Expert* Human Feedback!), would be expensive and time consuming, it might be the only reliable method to significantly improve the quality of this training phase.

In the context of RAG, , it must be remembered that  given their inability to understand language, LLMs must be provided with a pre-digested, unambiguous collection of legal rules.  If, however, the legal knowledgebase contains inaccuracies, they will be reflected in the output. Unexpectedly, the practical usability of LLMs may depend on the availability of high-quality legal knowledgebases – not on the number of parameters in a neural network.  This leads the debate about the capabilities of LLMs back to an "old" problem: the aforementioned knowledge acquisition and representation bottleneck, the difficulty of establishing and formalizing the legal rules. It becomes apparent that machine learning cannot succeed – at least not in the legal profession – without the assistance of knowledge engineering, the science of emulating expert knowledge.[307] Word prediction by itself is of limited use if it cannot be supported with accurate legal resources.

In the context of prompting, it must be remembered that  although prompts are written in natural language and may thus *appear* easy to write, crafting optimal prompts requires a unique combination of legal and technical skills that may remain beyond the reach of many less experienced lawyers and, more importantly, users without legal training. Effective prompting requires familiarity with the technical constraints of langue models, including their inability to process longer texts[308] and the importance of positioning more important information at the beginning or at the end of a given prompt.[309]    Moreover, prompt engineering requires not only systematic experimentation to  evaluate the effects of various prompt formulations on the generated output,[310] but also legal expertise to provide adequate instructions and to evaluate the adequacy of such output. The latter fact alone constitutes a great challenge for users who do not have legal training – how could they iteratively improve the prompt if they are unable to assess the generated output?[311]

---

[306] for insights how the labelling quality affect a model, see: Ilias Chalkidis, Manos Fergadiotis, Prodromos Malakasiotis, Nikolaos Aletras & Ion Androutsopoulos, *LEGAL-BERT: The Muppets Straight Out of Law School*, FINDINGS 2020 CONF. EMPIRICAL METHODS NLP 2898 (2020).

[307] S.L. KENDALL, M. GREEN, AN INTRODUCTION TO KNOWLEDGE ENGINEERING (2007)

## Conclusions

The use of LLMs in legal practice requires a cost-benefit assessment in terms of time and quality. When evaluating the speed with which an LLM generates its output, it is necessary to consider the time and human resources required for its verification. The resources required to verify the generated content must be weighed against the resources required to write such content in a traditional manner. LLMs may "reduce the cost it takes humans to bullshit to zero while not lowering the cost of producing truthful or accurate knowledge."[312] Given the risk of overreliance, the question is not whether LLMs can perform a particular task but whether they should be used to perform such a task.

The risk of hallucinations must not be underestimated. What would be an acceptable "hallucination rate"? How would it translate into cost savings? The unprecedented ability to generate fluent and superficially plausible text creates unprecedented dangers: users may not trust a system operating with 90% accuracy if 10% of the hallucinations are easy to spot. They may, however, trust systems operating with 80% accuracy if 20% of the hallucinations are hard to detect. While layers may debate whether a generated statement constitutes a hallucination or an unprecedented but possible approach to a legal problem, users without legal training will simply assume that it is correct – as long as the statement seems plausible and is written in perfect "legalese." The idea of using LLMs to democratize access to the law is thus inherently flawed. The risks are highest for those who could benefit from LLMs most – users without legal training.

---

[308] Aydar Bulatov et al., *Scaling Transformer to 1M tokens and beyond with RMT,* ARXIV (Feb. 6, 2024) https://arxiv.org/pdf/2304.11062.pdf; Weizhi Wang et al., *Augmenting Language Models with Long-Term Memor,* ARXIV y (Jun. 23, 2023) https://arxiv.org/pdf/2306.07174.pdf.

[309] Nelson F. Liu et al., *Lost in the Middle: How Language Models Use Long Contexts* (Nov. 20, 2023) https://arxiv.org/pdf/2307.03172.pdf; Tony Z. Zhao, Eric Wallace, Shi Feng, Dan Klein & Sameer Singh, *Calibrate Before Use: Improving Few-Shot Performance of Language Models,* PROC. 38TH INT'L CONF. MACH. LEARNING at 4 (2021) (demonstrating that content near the end of a prompt can significantly impact on a model's outputs).

[310] Zamfirescu-Pereira et al., *supra* note 235.

[311] Chris Draper and Nicky Gillibrand, *The Potential for Jurisdictional Challenges to AI or LLM Training Datasets*, PROC. ICAIL 2023 WORKSHOP ON ARTIFICIAL INTELLIGENCE FOR ACCESS TO JUSTICE; Dahl et al., *supra* note 8, at 14.

[312] Ezra Klein, (Jan. 6, 2023). The Ezra Klein Show. https://www.nytimes.com/2023/01/06/podcasts/transcript-ezra-klein-interviews-gary-marcus.html.

LLMs are the product of thousands of human choices - and whenever there is human choice there is room for error, subjectivity, incorrect interpretations or applications of the law, unclear instructions, and sloppy implementations. From the selection of training data, the drafting of annotation guidelines and the evaluation of model outputs, to the formulation of prompts and the creation of retrieval systems, the problem is always the same: all those tasks require legal expertise. And legal expertise cannot be replaced with word prediction alone.

The manner LLMs are used is largely shaped by the popular perceptions of what such models are and of what they can do – perceptions that are often based on a blind but uninformed belief in the transformative character of this technology. Apart from the fact that many lawyers underestimate the risk of hallucinations and are thus more likely to rely on the generated text without evaluating its correctness, the necessity to expand significant resources to verify each generated response largely defies the purpose of deploying LLMs: that of saving resources and optimizing human performance. It is open to debate whether the greatest danger lies in the models' propensity to hallucinate, in the users' ignorance of what LLMs were trained to do or in the user's lack of expertise when formulating prompts and evaluating model outputs.